\documentclass{article}

 \usepackage[preprint]{neurips_2026}

\usepackage[utf8]{inputenc} 
\usepackage[T1]{fontenc}    
\usepackage{hyperref}       
\usepackage{url}            
\usepackage{booktabs}       
\usepackage{amsfonts}       
\usepackage{nicefrac}       
\usepackage{microtype}      
\usepackage{xcolor}         

\usepackage{booktabs} 
\usepackage{kotex}
\usepackage{setspace}

\usepackage{rotating}

\usepackage{amsmath}
\usepackage{amssymb}
\usepackage{mathtools}
\usepackage{amsthm}
\usepackage{thmtools,thm-restate}
\usepackage{cancel}
\usepackage{pifont}

\usepackage{multicol,multirow}
\usepackage{subcaption}

\usepackage{float}
\usepackage{wrapfig}
\usepackage{thm-restate}
\usepackage{titletoc}
\usepackage[capitalize,noabbrev]{cleveref}

\title{One-Point Contraction: Erasing Representational Separability toward Irreversible Deep Forgetting}

\author{%
  Jaeheun Jung\\
  Department of Mathematics\\
  Korea University\\
  Seoul, Republic of Korea \\
  \texttt{wodsos@korea.ac.kr} \\
  \And
  Bosung Jung\\
  Department of Mathematics\\
  Korea University\\
  Seoul, Republic of Korea \\
  \texttt{2018160026@korea.ac.kr} \\
  \AND
  Suhyun Bae\\
  Department of Mathematics\\
  Korea University\\
  Seoul, Republic of Korea \\
  \texttt{baeshstar@korea.ac.kr} \\  
  \And
  Donghun Lee\thanks{corresponding author}\\
  Department of Mathematics\\
  Korea University\\
  Seoul, Republic of Korea \\
  \texttt{holy@korea.ac.kr} \\
}

\newcommand{\OPC}{\textsc{OPC}}
\newcommand{\FMR}{\textsc{FM}-recovery}

\begin{document}

\maketitle

\begin{abstract}
Machine unlearning is usually evaluated by what the classifier outputs: forget-set accuracy, confidence, membership-inference scores. 
We show that this is not enough. Across 14 representative unlearning methods on
CIFAR-10 and SVHN, a single linear map fitted on a held-out calibration set, with no access to the forgotten data, reverses the unlearning in seconds and recovers forget-set accuracy to within a few percent of the original model. 
Recovered features even support pixel-level reconstruction through a generic decoder. 
We call this diagnostic \emph{Feature Mapping Recovery} (\FMR{}). 
The pattern it exposes is uniform: current unlearning methods do not erase information from the
representation, they apply an invertible linear distortion that hides it from one particular prediction head. 
We propose \emph{One-Point Contraction} (\OPC{}), an unlearning objective that collapses forget-set features to the origin while leaving the retain-set geometry intact. 
We prove that this contraction is equivalent to driving the predictive distribution to maximum entropy, so the same mechanism delivers behavioral forgetting and representation-level erasure at once.
Forgotten queries land in a region the network treats as out-of-distribution, and the gradient signal on those queries collapses along with their features. 
\OPC{} is the only method in our benchmark that survives \FMR{}, resists relearning and gradient-inversion attacks, and decouples forget from retain features in entangled settings, all without sacrificing retain or test accuracy.

\end{abstract}

\vspace{-2mm}
\section{Introduction}
\vspace{-3mm}

When a model is tasked to forget, what exactly should disappear?
The standard answer is operational: produce a model that, on the data it was asked to forget, behaves as if it had never seen them.
Modern machine learning systems are routinely trained on data that later must be removed—copyrighted content, personally identifiable information, or mislabeled examples subject to "right-to-be-forgotten" requests \cite{Ginart2019-by, Bourtoule2021-mf, Nguyen2025-mj}.
Because retraining foundation-scale models from scratch is rarely feasible, the field has converged on \emph{approximate} machine unlearning (MU): lightweight parameter updates that aim to match the behavior of a from-scratch retrain on a target forget set \cite{Golatkar_2020_CVPR, GA, SCRUB_NegGrad+, salun}.

The behavioral framing has been productive, but it has also fixed the yardstick.
Because retraining is too expensive to use as a routine reference, MU methods are compared on output-level surrogates: forget-set accuracy, confidence, entropy, membership-inference success.
These metrics ask whether the final classifier still acts as if it remembers.
They do not ask whether the model still encodes what it was told to forget.

That distinction is not academic.
Consider an unlearned model that returns near-uniform softmax probabilities on every forget-set query—the canonical signature of successful forgetting.
If the penultimate features of those queries still cluster by their original labels, the classifier is merely "confused" while the representation remains intact.
A downstream actor with white-box access needs only a simple linear probe to recover the forgotten information.
In such cases, the knowledge has not vanished; it has merely been hidden from the original prediction head.

\begin{figure*}[t]
    \centering
    \begin{subfigure}[t]{0.29\textwidth}
        \includegraphics[width=\textwidth]{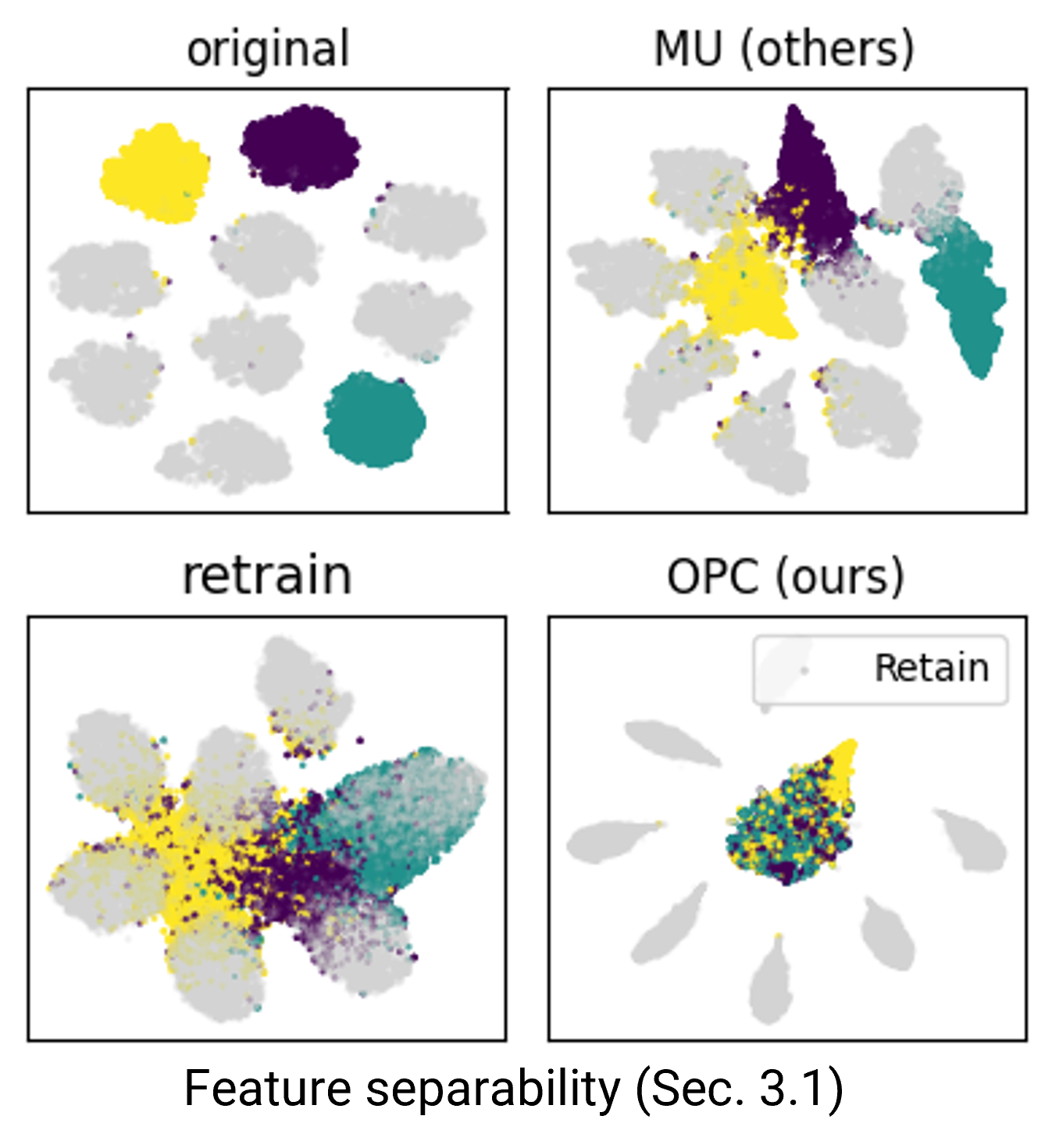}
        \caption{tSNE visualization}
        \label{subfig:concept_separability}
    \end{subfigure}
    \begin{subfigure}[t]{0.53\textwidth}
        \includegraphics[width=\textwidth]{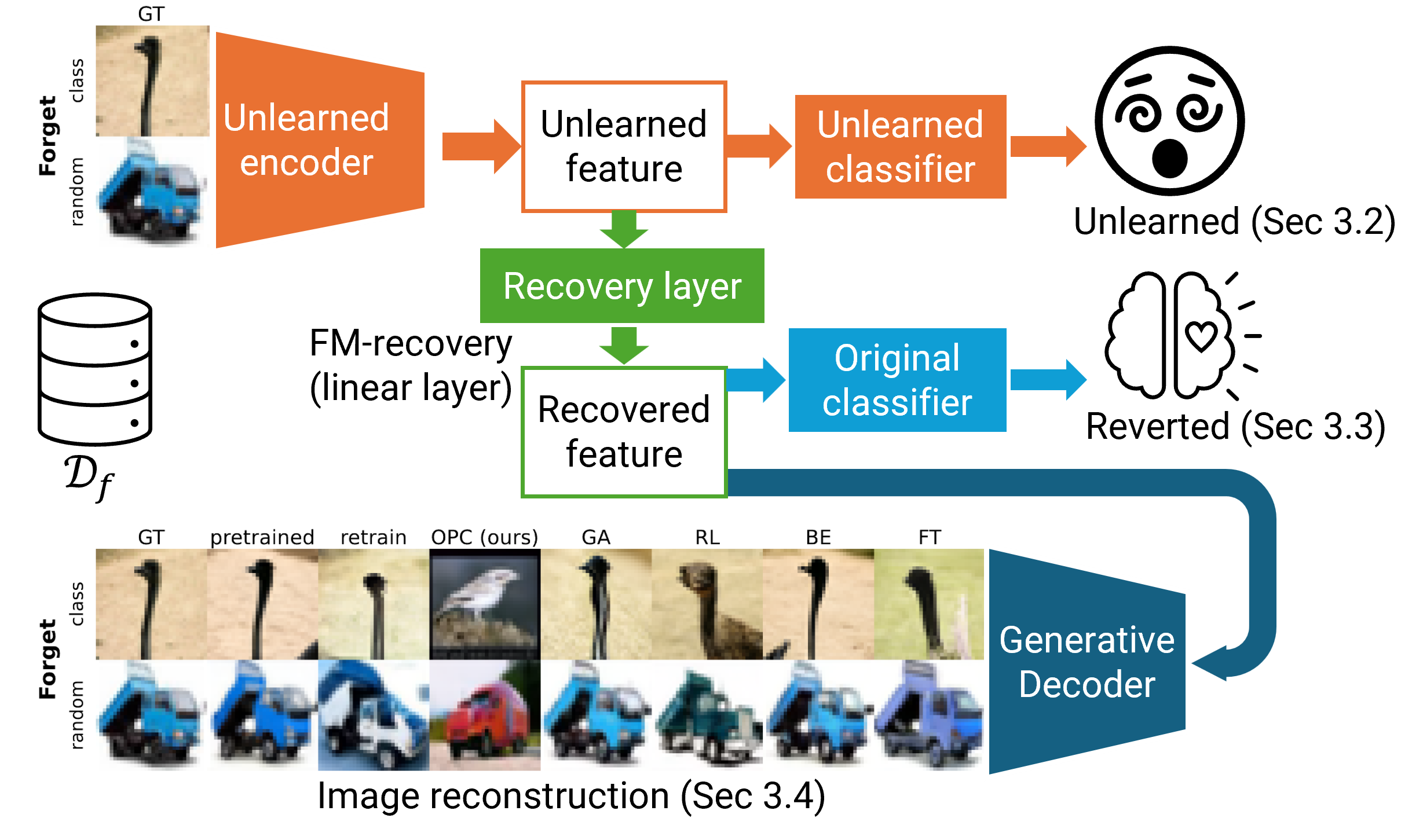}
        \caption{Concept of MU reverting}
        \label{subfig:concept_recovery}
    \end{subfigure}
    \begin{subfigure}[t]{0.16\textwidth}
        \includegraphics[width=\textwidth]{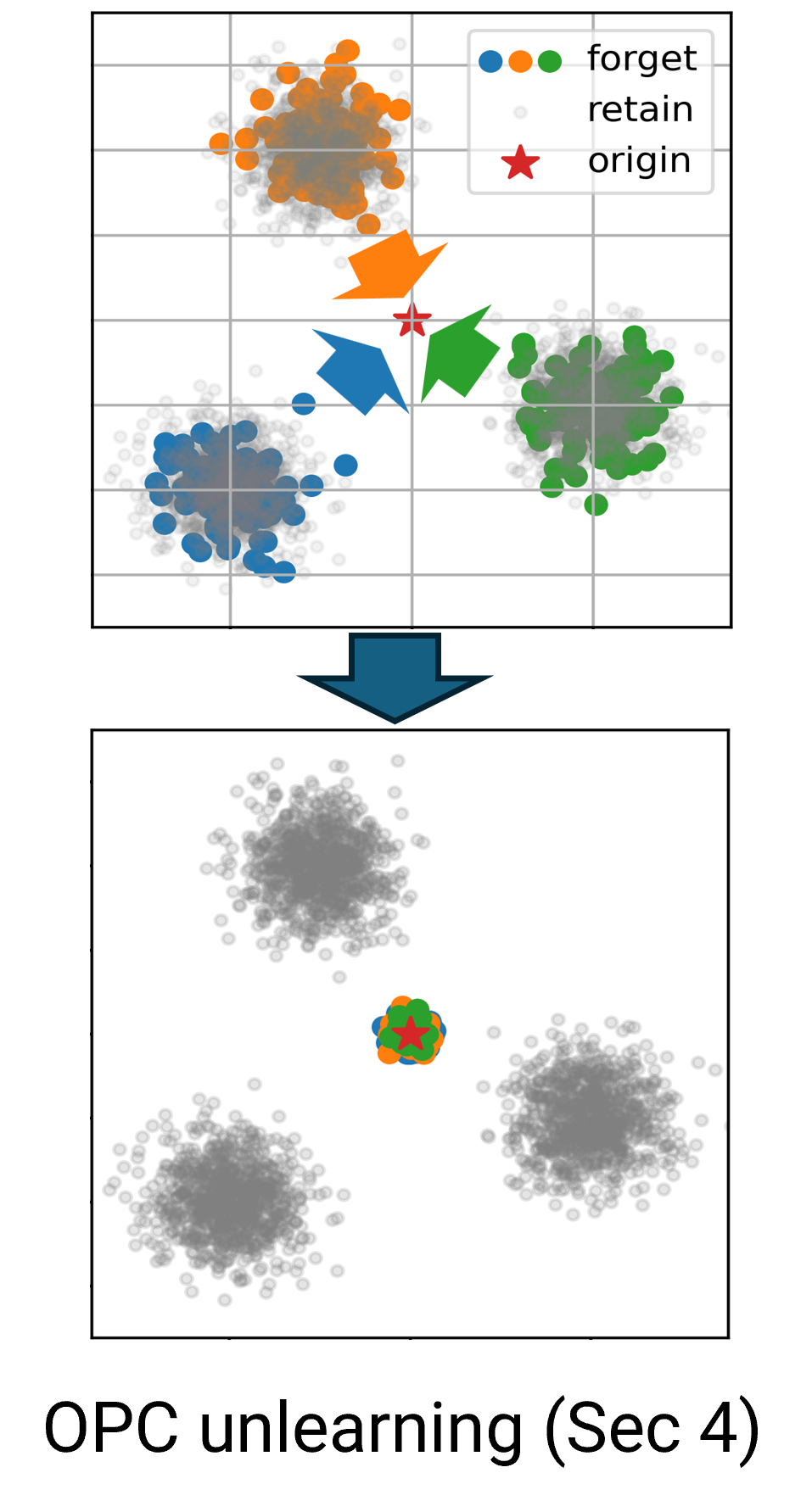}
        \caption{Concept of OPC}
        \label{subfig:concept_OPC}
    \end{subfigure}
    \vspace{-1mm}
    \caption{Visualization of concepts.
    \cref{subfig:concept_separability} shows that separability among forget features survives machine unlearning.
    \cref{subfig:concept_recovery} depicts the pipeline of \cref{sec:recovery}, which evaluates the depth of forgetting by reverting the MU process.
    \cref{subfig:concept_OPC} illustrates \OPC{}, our proposed MU method, which renders forget features indistinguishable.}
    \vspace{-5mm}
\end{figure*}

We call this failure mode \textbf{shallow forgetting}, and we argue it is the typical outcome of current MU rather than a rare pathology.
To address this, we introduce a diagnostic framework to detect shallow forgetting and a novel unlearning paradigm that is inherently resilient to it.
Our contributions are three-fold:

\textbf{Shallow forgetting is the majority, not the exception (\cref{sec:recovery}).}
We introduce \emph{Feature Mapping Recovery} (\FMR{}), a minimalist diagnostic that fits a linear map from the unlearned model’s penultimate features back to the original feature space. Using only a small calibration set disjoint from the forget-set, FM-recovery restores forget accuracy to within 5--15\% of the original model across 14 representative MU methods. This reveals that current methods produce an invertible linear distortion of an otherwise intact feature space.

\textbf{One-Point Contraction (\cref{sec:OPC}).}
We propose an unlearning objective that does not just hide forget-set information from the head, it removes it from the representation.
\OPC{} contracts every forget-set feature toward a single fixed point while preserving retain-set geometry, and choosing that point to be the origin reduces the whole procedure to a localized norm penalty on the logits, making provable explosion of uncertainty on forget-set.
The resulting objective is one extra term in a standard fine-tuning loop and a few lines of code.
Because the contraction acts in the penultimate feature space rather than at the output, no relinearization of the head can recover what it removes.

\textbf{Efficacy-Prioritized Unlearning.}
We introduce a fresh perspective that prioritizes absolute forgetting efficacy over the traditional goal of strictly mimicking a retrained model. 
While conventional MU often fails to induce true malfunctionality—especially when tasked with erasing random, non-homogeneous forget queries—our approach ensures that forgotten data is rendered fundamentally inert. 
This ensures consistent and robust erasure across diverse scenarios, addressing the practical complexities of real-world "right-to-be-forgotten" requirements.

\vspace{-2mm}
\section{Related Works}\label{sec:Relworks}
\vspace{-3mm}

\paragraph{Machine unlearning.}
Machine unlearning (MU) aims to remove the influence of a specific forget-set $\mathcal{D}_f$ from a trained model while maintaining performance on a retain-set $\mathcal{D}_r$, motivated by regulatory compliance (e.g., GDPR's ``right to be forgotten'') \cite{GDPR16} and practical demands such as excising poisoned, copyrighted, or outdated data \cite{ShafahiHNSSDG18,henderson2023foundation,schoepf2024potion}.
Exact unlearning via retraining from scratch is prohibitively expensive for modern models; the SISA framework \cite{Bourtoule2021-mf} mitigates this by partitioning training data so that only affected shards need retraining, while certified deletion methods \cite{guo2019certified,sekhari2021remember} provide formal guarantees but are restricted to convex or simple models. 
Approximate MU methods have since emerged as the practical workhorse, spanning gradient-ascent approaches \cite{GA,Golatkar_2020_CVPR}, gradient-descent variants \cite{neel2021deletion,chien2024langevin}, teacher--student distillation \cite{bad_teacher}, weight saliency \cite{salun,l1sparse/MIAp}, synaptic dampening \cite{SSD,LFSSD}, boundary shifting \cite{BEBS}, fine-tuning approaches \cite{FT}, and Nash bargaining formulations \cite{munba}. 
Collectively, these methods achieve competitive scores on standard logit-based metrics such as forget accuracy (FA), retain accuracy (RA), and membership inference attack (MIA) success~\cite{MIA,carlini2021membership}.
In this paper, we benchmark 14 representative methods listed in \cref{appendix:methods} and demonstrate that, despite strong output-level scores, they primarily manipulate model outputs without fundamentally erasing internal representations.

\vspace{-1mm} 
\paragraph{Shallow forgetting and relearning attacks.}
The shallowness of MU has recently been investigated from multiple angles.
Relearning attacks~\cite{ha2025unlearning,huunlearning,deeb2024unlearning,relearning} fine-tune the unlearned model on a small subset of forget data and observe rapid performance recovery, demonstrating that supposedly erased knowledge persists in the weights.
This vulnerability has been confirmed in both vision~\cite{ha2025unlearning} and LLM settings~\cite{huunlearning,deeb2024unlearning,xu2025unlearning}, where unlearned models can be ``jogged'' back to their original behavior with minimal data.
\cite{hayes2024inexact} further argue that inexact unlearning methods require significantly more careful evaluation to avoid a false sense of privacy, while \cite{RUM} analyze what makes unlearning hard and attribute the difficulty to the entanglement between forget and retain representations. 
At the representation level, CKA similarity \cite{CKA1,CKA2} has been used to diagnose residual information: \cite{kim2026we} compare unlearned features against retrained and original models, finding persistent structural similarity, and \cite{ha2025unlearning} use layer-wise CKA to quantify how little internal representations change after unlearning.

\vspace{-1mm} 
\paragraph{Reconstruction and privacy attacks on MU.}
Beyond relearning, reconstruction attacks directly exploit the residual information in unlearned models.
\cite{hu2024_reconattack} show that gradient-inversion techniques applied to the parameter difference between original and unlearned models can reconstruct forget-set images, while \cite{bertran2024_linear_recon_attack} demonstrate that even simple linear models are vulnerable to reconstruction after unlearning.
Membership inference attacks \cite{MIA,carlini2021membership} further test whether unlearned samples remain distinguishable from unseen data.
Adversarial evaluations~\cite{goel2022towards} and probabilistic verification~\cite{sommer2022athena} have been proposed to provide stronger guarantees, yet these focus on output-level behavior and do not inspect the feature geometry. 
Our work complements these threat models by exposing vulnerability at the \emph{representation} level: even when logit-based metrics indicate successful unlearning, the underlying features preserve enough structure for both classification recovery and image reconstruction through external decoders.

\begin{figure*}[t]
\centering
    \includegraphics[width=0.98\textwidth]{plots/plots_tSNE/tsne_comparison_cifar10cls30_feature.png}
    \caption{tSNE visualization of unlearned features on CIFAR10 30\% class unlearning scenario. The gray-colored points represent retain features and forget features were colored along its class labels.}
    \label{fig:tsne-feature-class-cifar10}
    \vspace{-0.6cm}
\end{figure*}

\vspace{-2mm}
\section{Undo MU: Feature separability matters}\label{sec:recovery}
\vspace{-3mm}

Investigating the depth of forgetting in unlearned features, we found that most MU methods induce only linear distortions in the representation space and the unlearned forget features are still linearly separable along their true class labels. 
Furthermore, this linear distortion is invertible; thus, we could \textbf{undo} the MU process and recover the original (pretrained) feature by adding a single linear layer, depicted in \cref{subfig:concept_recovery}.

In this section, we conduct empirical experiments to support the above claim. Specifically, we apply 14 baseline MU methods on CIFAR10 and SVHN under a class 30\% unlearning scenario and a random 10\% unlearning scenario. 
The results on CIFAR10 with the class unlearning scenario are focused on in this section, and other results can be found in \cref{appendix:more-experiments}.
In \cref{subsec:recovery-tsne}, we demonstrate feature separability of the forget features. 
Although all baselines show fine unlearning performance (\cref{subsec:recovery-performance}), it was possible to revert the unlearned model to the original model through feature-mapping recovery, which we explain in \cref{subsec:recovery-FMrecovery}. 
Beyond classification accuracy, we show in \cref{subsec:recovery-decoder} the quality of recovered features is sufficient to reconstruct the original images, using a decoder trained to generate the original image from the original forget features.

\vspace{-1mm}
\subsection{Separability of unlearned feature}\label{subsec:recovery-tsne}
\vspace{-1.5mm}

Our work was motivated from the observation of unlearned features and their separability, which we visualized in \cref{fig:tsne-feature-class-cifar10}.
Although MU were performed, we observed that the features still form clusters along their true class label, similar to original (pretrained) model. 

Based on this observation, we hypothesize linear separability persists in the representation space, and that there exists a linear transformation between unlearned features and pretrained features; such a transformation should not exist if ideal forgetting were performed. We investigate the existence of such a linear transformation in \cref{subsec:recovery-FMrecovery} and \cref{subsec:recovery-decoder}.

\vspace{-1mm}
\subsection{MU Performance}\label{subsec:recovery-performance}
\vspace{-1.5mm}

\begin{table*}[b]
\caption{Unlearning performance on 30\% Class unlearning scenario. }
\centering
\resizebox{0.49\linewidth}{!}{%
\begin{tabular}{cccccc}
\toprule
\textbf{CIFAR10} & Train $\mathcal{D}_f(\downarrow)$ & Train $\mathcal{D}_r(\uparrow)$ & Test $\mathcal{D}_f(\downarrow)$ & Test $\mathcal{D}_r(\uparrow)$ & $\mathbf{MIA}^e(\uparrow)$ \\
\midrule
Pretrained (Original) & 99.444 & 99.416 & 94.800 & 94.400 & 0.015 \\
Retrain & 0.000 & 99.981 & 0.000 & 91.700 & 1.000 \\
\textbf{OPC} (ours) & 0.000 & 99.606 & 0.000 & 93.143 & 1.000 \\
\midrule
GA \citep{GA} & 0.148 & 87.771 & 0.033 & 84.057 & 0.998 \\
RL \citep{RL} & 0.000 & 99.060 & 0.000 & 93.529 & 1.000 \\
BE \citep{BEBS} & 0.037 & 93.168 & 0.000 & 85.214 & 0.998 \\
FT \citep{FT} & 0.000 & 98.994 & 0.000 & 93.457 & 1.000 \\
NGD \citep{NGD} & 0.000 & 98.498 & 0.000 & 93.071 & 1.000 \\
NegGrad+ \citep{SCRUB_NegGrad+} & 0.000 & 98.638 & 0.000 & 93.014 & 1.000 \\
EUk \citep{EUk/CFk} & 0.000 & 99.616 & 0.000 & 94.629 & 1.000 \\
CFk \citep{EUk/CFk} & 0.170 & 99.759 & 0.167 & 94.929 & 1.000 \\
SalUn \citep{salun} & 0.000 & 99.743 & 0.000 & 94.786 & 1.000 \\
SCRUB \citep{SCRUB_NegGrad+} & 0.000 & 98.060 & 0.000 & 93.457 & 1.000 \\
BT \citep{bad_teacher} & 8.578 & 99.502 & 7.533 & 95.286 & 1.000 \\
$l1$-sparse \citep{l1sparse/MIAp} & 0.000 & 99.425 & 0.000 & 94.386 & 1.000 \\
MUNBa \citep{munba} & 0.222 & 99.467 & 0.367 & 94.729 & 1.000 \\
DELETE \citep{delete} & 0.007 & 99.505 & 0.367 & 95.229 & 1.000 \\
\bottomrule
\end{tabular}%
}
\resizebox{0.49\linewidth}{!}{%
\begin{tabular}{cccccc}
\toprule
\textbf{SVHN} & Train $\mathcal{D}_f(\downarrow)$ & Train $\mathcal{D}_r(\uparrow)$ & Test $\mathcal{D}_f(\downarrow)$ & Test $\mathcal{D}_r(\uparrow)$ & $\mathbf{MIA}^e(\uparrow)$ \\
\midrule
Pretrained (Original) & 99.531 & 99.172 & 94.960 & 91.110 & 0.009 \\
Retrain & 0.000 & 99.997 & 0.000 & 92.440 & 1.000 \\
\textbf{OPC} (ours) & 0.011 & 99.612 & 0.009 & 94.142 & 1.000 \\
\midrule
GA \citep{GA} & 73.220 & 96.477 & 62.618 & 86.270 & 0.381 \\
RL \citep{RL} & 0.000 & 99.997 & 0.000 & 93.876 & 1.000 \\
BE \citep{BEBS} & 1.240 & 95.355 & 0.910 & 78.690 & 0.990 \\
FT \citep{FT} & 0.034 & 99.997 & 0.009 & 94.535 & 1.000 \\
NGD \citep{NGD} & 0.000 & 99.997 & 0.000 & 94.854 & 1.000 \\
NegGrad+ \citep{SCRUB_NegGrad+} & 0.000 & 97.997 & 0.000 & 91.642 & 1.000 \\
EUk \citep{EUk/CFk} & 0.000 & 99.997 & 0.000 & 92.826 & 1.000 \\
CFk \citep{EUk/CFk} & 0.000 & 99.997 & 0.000 & 92.945 & 1.000 \\
SalUn \citep{salun} & 0.000 & 99.990 & 0.000 & 93.910 & 1.000 \\
SCRUB \citep{SCRUB_NegGrad+} & 0.008 & 94.995 & 0.000 & 89.129 & 1.000 \\
BT \citep{bad_teacher} & 8.633 & 99.210 & 4.904 & 93.437 & 1.000 \\
$l1$-sparse \citep{l1sparse/MIAp} & 0.000 & 98.954 & 0.000 & 92.872 & 1.000 \\
MUNBa \citep{munba}& 0.038 & 99.901 & 0.009 & 93.398 & 1.000 \\
DELETE \citep{delete} & 0.011 & 99.341 & 0.400 & 92.899 & 1.000 \\
\bottomrule
\end{tabular}%
}
\label{tab:cifar10-cls30-perf}
\end{table*}

We list the logit-based performance of each benchmarked MU method in \cref{tab:cifar10-cls30-perf}. All MU methods show good results: the forget accuracy is nearly 0\%, the retain accuracy (RA) and test accuracy (TA) are preserved. The $MIA^e$ metric (MIA efficacy, ~\cite{MIA}) also supports that all baselines achieved proper unlearning, as evidenced by good MU scores. In the next section, we revert each MU process and show how these scores change.

\vspace{-1mm}
\subsection{FM-recovery: Undo MU via Feature Mapping}\label{subsec:recovery-FMrecovery}
\vspace{-1.5mm}

\begin{figure*}[h]
    \includegraphics[width=\linewidth]{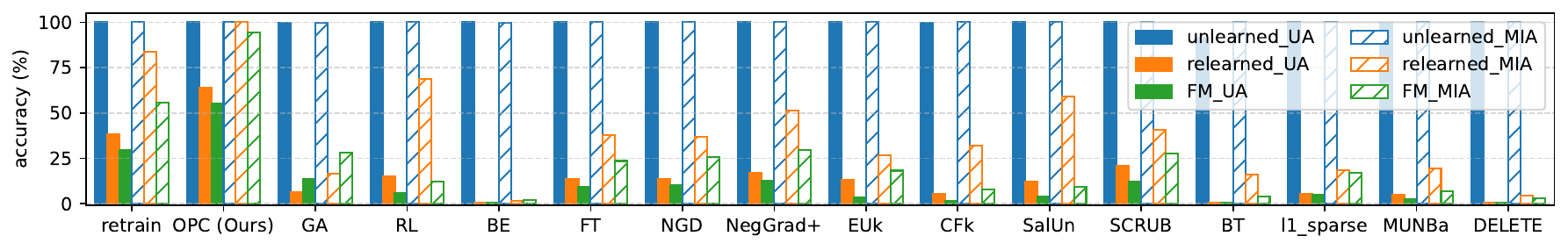}
    \vspace{-0.6cm}
    \caption{Effect of Relearning (orange) and FM recovery (green) compared to unlearned model (blue), in UA (100\%-FA) and $MIA^e$ for CIFAR10 class 30\% unlearning scenario. }
    \label{fig:recovery_UA}
    \vspace{-0.4cm}
\end{figure*}

From the observation of feature separability in \cref{subsec:recovery-tsne}, we hypothesize the existence of a linear relationship between original forget features and unlearned forget features. To discover this linear transformation, we consider the following least squares problem:

\vspace{-1mm}
\begin{equation}\label{eqn:lstsq-featuremap}
    W^*=\underset{W}{\arg\min} \sum_{x\in\mathcal{D}} \|f_{\theta^0}(x)-Wf_{\theta^{un}}(x)\|_2^2 ,
\end{equation}
\vspace{-1mm}

where $f_\theta$ is the feature encoder of the model, $\theta^0$ is the parameter of original model, $\theta^{un}$ is the parameter of unlearned model and $\mathcal{D}$ is the calibration dataset, which is disjoint from the training dataset. We set $\mathcal{D}$ using the validation set for the experiment. Since the least squares problem has a closed form solution, the feature mapping matrix $W^*$ can be computed quickly, taking less than 4 seconds in our environment.

After obtaining $W^*$, we can recover the unlearned features by simply multiplying by $W^*$; we call this approach feature mapping recovery or simply FM-recovery. After applying FM-recovery, we use the original model's classifier head to make model predictions and evaluate performance. For the comparison, we also report the relearning\cite{relearning} attack results.

The results are visualized in \cref{fig:recovery_UA} and listed in \cref{tab:cifar10-cls30-featuremap}. 
Surprisingly, FM-recovery was able to recover the original model's performance on the forget dataset almost perfectly, usually more than the relearning, except for retrain and OPC (ours). 
Also, our FM-recovery acts as a efficient proxy for the relearning: recovery of relearning and FM-recovery are consistent.

Since both the classifier head and FM-recovery layer are linear, their composition is still linear; thus, this recovery of unlearned models indicates the linear separability of the features. 
We further validate this linear separability in \cref{subsec:exp-headrecovery} by directly finding the classifier head.

\vspace{-1mm}
\subsection{Qualitative evaluation via generative decoder}\label{subsec:recovery-decoder}
\vspace{-1.5mm}

\begin{figure*}[t]
  \centering
\begin{subfigure}[t]{0.99\linewidth}
      \includegraphics[width=\textwidth]{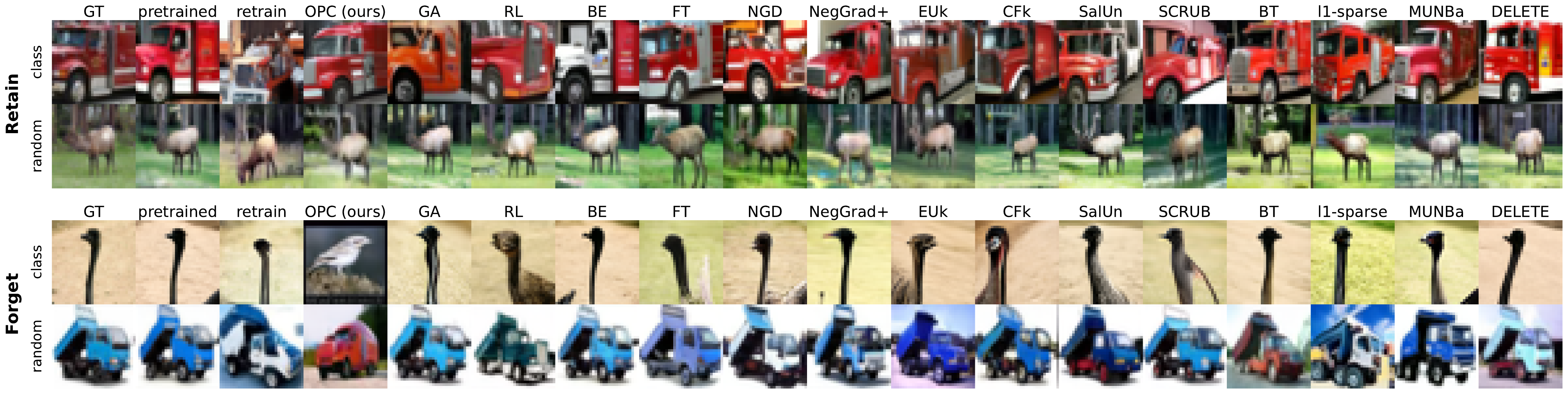}
\end{subfigure} 
\caption{Reconstruction from recovered features. The target images are sampled from $\mathcal{D}_r$ and $\mathcal{D}_f$ of CIFAR10 under both the 30\% class and the 10\% random unlearning scenarios. GT represents the ground truth image from the dataset and others are the results of reconstruction from each model.}
\vspace{-6mm}
\label{fig:recovery-recon}
\end{figure*}

Although FM-recovery with the original classifier head successfully recovered the original performance, inter-class recovery remains questionable, as features encode more details about the input image. 
To qualitatively investigate this, we train an external generative decoder that is trained on the training dataset to reconstruct the input image $x$ from the original feature $f_{\theta^0}(x)$. In implementation, we trained DDPM~\cite{DDPM} and generated the image from recovered features.

\cref{fig:recovery-recon} shows the reconstruction results for retain data and forget data. For retain data, all recovered features except those from the retrained model were similar enough to the original features, allowing the decoder to reconstruct the images to be almost identical to the ground truth. 
The features recovered from the retrained model were somewhat different in terms of car shape or the deer's behavior, but following  the behavior of the original model is considered more desirable in MU: the model should behave identically on the retain-set $\mathcal{D}_r$.

For the forget data, the recovered features generated almost identical images through the generative decoder with all details of the images (e.g., bird type, background color, car type and car direction) being restored, except for OPC (ours, \cref{sec:OPC}) and partially for retraining.

\vspace{-1mm}
\subsection{CKA analysis}\label{subsec:recovery-CKA}
\vspace{-1.5mm}

\begin{wrapfigure}{rb}{8cm}
  \centering
  \vspace{-6mm}
  \includegraphics[width=\linewidth]{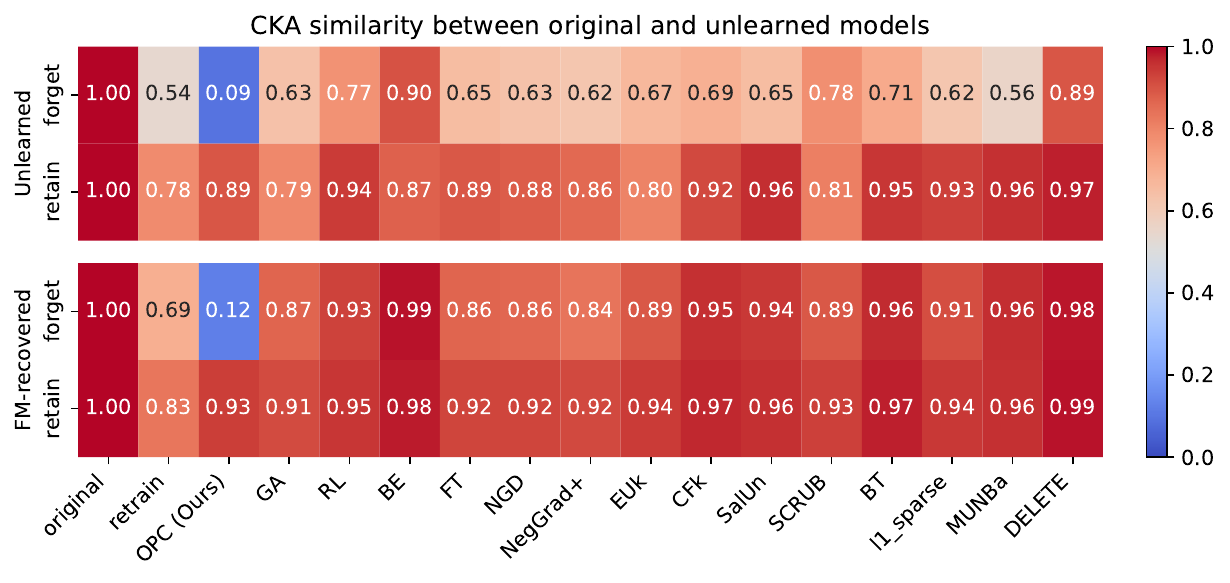}
    \caption{CKA similarity between original features and unlearned features, measured on forget-set $\mathcal{D}_f$ and retain-set $\mathcal{D}_r$ of CIFAR10 class unlearning, before and after FM-recovery.}
    \label{fig:recovery-cifar10cls30-CKA}
  \vspace{-4mm}
\label{fig:CKA_cifar10_pretrained}
\end{wrapfigure}

In \cref{fig:recovery-cifar10cls30-CKA}, we show the CKA similarity between original features and unlearned features, and how FM-recovery affects them. 
Before FM-recovery, unlearned models produce forget features with a moderate gap compared to the original features, while preserving the retain features. 
However, FM-recovery reverts unlearned models to produce forget features close to the original features as CKA similarity is measured to be high, while retraining shows resistance; the recovered features are not as similar to the original features. 
Although CKA is invariant under isotropic scaling and orthogonal transformation, the CKA measurement also fails to capture the reversibility of forgetting in MU.

FM-recovery demonstrates that most MU methods merely induce an invertible linear distortion, leaving forget-set features highly informative and vulnerable to reconstruction without additional training. 
Even retraining is not an absolute safeguard due to generalization. 
Collectively, this proves current unlearning is \emph{shallow} and fundamentally reversible, necessitating MU to be robust with feature-level erasure.

\vspace{-2mm}
\section{OPC: One-Point Contraction}\label{sec:OPC}
\vspace{-3mm}

The linear separability of forget unlearned features makes MU shallow and allows \FMR{} to recover the forget features and performance; while retraining were resistant. 
Hence, natural desiderata of deeper MU method arises and we answer the call by proposing a simple yet novel MU algorithm called One-Point Contraction (\textbf{OPC}), which collapses the forget feature toward origin to remove the separability. 

In this section, we introduce OPC unlearning, and investigate the separation between forget features and retain features on feature space. 
This is to challenge the feature entanglement problem: a long-lasting difficulty in MU, in which the forget set must be erased even when its elements are entangled with the retain set.

OPC shows remarkably high efficacy on entangled scenario; we show the MU results on random unlearning and half-class unlearning which considers forget dataset consist of random samples along all classes and half of the forget classes (the rest half would be the retain data, denoted by $\mathcal{D}_r^e$).

\vspace{-1mm}
\subsection{Method: OPC to break distinguishability}\label{subsec:OPC-method}
\vspace{-1.5mm}

Motivated by the intuition that the linear separability of forget features makes unlearned model reversible, we propose to push the predictions on forget data toward a single point, the origin, and make the forget features indistinguishable.

We implement the contraction as an optimization problem to minimize the $L_2$ norm of the prediction logits $\mathbf{m}_\theta(x)$ for the forget samples $x \in \mathcal{D}_f$, where $\mathbf{m}_\theta$ is the composition of the encoder $f_\theta$ and the classifier head. To preserve retain performance, we also minimize an objective loss $\mathcal{L}$ on retain-set $\mathcal{D}_r$. The loss function of OPC for classification can be written as:

\begin{equation}\label{eq:opc_loss_ce}
\begin{aligned}
     \mathcal{L}_{OPC} = \lambda_r \mathbb{E}_{x,y\sim\mathcal{D}_r} \mathcal{L}(\mathbf{m}_\theta(x),y)+
     \lambda_f\mathbb{E}_{x\sim\mathcal{D}_f}\|\mathbf{m}_\theta(x)\|_2 ,
\end{aligned}
\end{equation}
where $\lambda_r$ and $\lambda_f$ are balancing hyperparameters.

The minimization of the logit norm can induce both an explosion of uncertainty and a minimization of the feature norm $\|f_\theta(x)\|$. 
We provide rigorous mathematical justification in \cref{appendix:theory}. Specifically, we propose a sharp lower bound for Shannon entropy with respect to the logit norm. We also provide gradient analysis showing that feature norm reduction occurs while avoiding the trivial solution that modifies only the classifier head.

\vspace{-1mm}
\subsection{Challenge against Entanglement}\label{subsec:OPC-tsne}
\vspace{-1.5mm}

\begin{figure*}[t]
\centering
    \includegraphics[width=0.98\textwidth]{plots/plots_tSNE/tsne_comparison_svhnele10_feature.png}
    \caption{tSNE visualization of unlearned features in the SVHN random 10\% unlearning scenario. The gray-colored points represent retain features and forget features are colored by their class labels.}
    \vspace{-2mm} 
    \label{fig:tsne-feature-rand-svhn}
\end{figure*}

\begin{figure*}[t]
\centering
    \includegraphics[width=0.98\textwidth]{plots/plots_tSNE/tsne_comparison_svhncls30ele50_feature.png}
    \caption{tSNE visualization of unlearned features in the CIFAR10 half-class 30\% unlearning scenario. The gray-colored points represent features from $\mathcal{D}_r^r$. Features from $\mathcal{D}_f$ and $\mathcal{D}_r^e$ are colored w.r.t their class labels.}
    \vspace{-4mm}
    \label{fig:tsne-feature-halfcalss-svhn}
\end{figure*}

Regarding the MU requirement that the model should work well on retain set, detaching the forget features from the retain features of the same class is mandatory for effective MU; if the forget features are highly entangled to retain features, any linear classifier that achieves strong forgetting with high UA will invoke catastrophic forgetting: the retain features cannot be mapped to the correct prediction since they are entangled with the forget features.

In \cref{fig:tsne-feature-rand-svhn} and \cref{fig:tsne-feature-halfcalss-svhn}, almost all MU methods including retraining struggle to separate the features due to over-generalization unlike the unentangled scenario in \cref{fig:tsne-feature-class-cifar10}. As a result, almost all of them fails to reduce model performance on $\mathcal{D}_f$ or invoke catastrophic forgetting on $\mathcal{D}_r$. 

However, OPC clearly separates the forget features from retain feature distribution and break the linear separability among forget features on all unlearning scenario, achieving both high MU efficacy (\cref{subsec:OPC-performance}) and resistance to \FMR{} (\cref{subsec:OPC-FMR}).

\vspace{-1mm}
\subsection{MU performance of OPC}\label{subsec:OPC-performance}
\vspace{-1.5mm}

\begin{table*}[t]
\caption{Unlearning performance on 10\% random unlearning scenario}
\centering
\resizebox{0.48\linewidth}{!}{%
\begin{tabular}{cccccc}
\toprule
\textbf{CIFAR10} & $\mathcal{D}_f (\downarrow)$ & $\mathcal{D}_r(\uparrow)$ & $\mathcal{D}_{test}(\uparrow)$ & $\mathbf{MIA}^e(\uparrow)$ &$\mathbf{MIA}^p(\downarrow)$ \\
\midrule
Pretrained (Original)& 99.356 & 99.432 & 94.520 & 0.015 & 0.545 \\
Retrain & 90.756 & 99.995 & 90.480 & 0.149 & 0.577 \\
\textbf{OPC} (ours) & 8.267 & 99.993 & 92.920 & 1.000 & 0.625 \\    
\midrule
GA\citep{GA} & 99.267 & 99.435 & 94.340 & 0.018 & 0.544 \\
RL\citep{RL} & 93.356 & 99.948 & 93.680 & 0.272 & 0.570 \\
BE\citep{BEBS} & 99.378 & 99.440 & 94.480 & 0.016 & 0.545 \\
FT\citep{FT} & 95.267 & 99.694 & 92.890 & 0.082 & 0.548 \\
NGD\citep{NGD} & 95.133 & 99.654 & 93.280 & 0.081 & 0.544 \\
NegGrad+\citep{SCRUB_NegGrad+} & 95.578 & 99.731 & 93.300 & 0.082 & 0.549 \\
EUk\citep{EUk/CFk} & 99.044 & 99.854 & 93.670 & 0.017 & 0.540 \\
CFk\citep{EUk/CFk} & 99.244 & 99.943 & 93.980 & 0.016 & 0.540 \\
SalUn\citep{salun} & 93.444 & 99.931 & 93.830 & 0.280 & 0.570 \\
SCRUB\citep{SCRUB_NegGrad+} & 99.222 & 99.511 & 94.060 & 0.047 & 0.548 \\
BT\citep{bad_teacher} & 91.422 & 99.341 & 93.010 & 0.560 & 0.558 \\
$l1$-sparse\citep{l1sparse/MIAp} & 92.889 & 97.360 & 90.980 & 0.129 & 0.539 \\
MUNBa \citep{munba} & 94.222 & 98.410 & 92.100 & 0.124 & 0.545 \\
DELETE \citep{delete} & 99.000 & 99.180 & 93.370 & 0.094 & 0.544 \\
\bottomrule
\end{tabular}%
}
\resizebox{0.48\linewidth}{!}{%
\begin{tabular}{cccccc}
\toprule
\textbf{SVHN} & $\mathcal{D}_f (\downarrow)$ & $\mathcal{D}_r(\uparrow)$ & $\mathcal{D}_{test}(\uparrow)$ & $\mathbf{MIA}^e(\uparrow)$ &$\mathbf{MIA}^p(\downarrow)$ \\
\midrule
Pretrained (Original)& 99.151 & 99.334 & 92.736 & 0.015 & 0.563 \\
Retrain & 92.947 & 99.998 & 92.490 & 0.154 & 0.583 \\
\textbf{OPC} (ours) & 7.493 & 99.949 & 92.636 & 1.000 & 0.607 \\
\midrule
GA\citep{GA} & 98.832 & 99.280 & 92.190 & 0.016 & 0.564 \\
RL\citep{RL} & 92.492 & 97.075 & 92.002 & 0.227 & 0.534 \\
BE\citep{BEBS} & 99.029 & 99.134 & 90.854 & 0.029 & 0.580 \\
FT\citep{FT} & 94.267 & 99.998 & 94.403 & 0.107 & 0.553 \\
NGD\citep{NGD} & 94.494 & 99.998 & 94.695 & 0.099 & 0.550 \\
NegGrad+\citep{SCRUB_NegGrad+} & 94.115 & 99.998 & 94.173 & 0.113 & 0.565 \\
EUk\citep{EUk/CFk} & 98.134 & 99.998 & 92.248 & 0.061 & 0.573 \\
CFk\citep{EUk/CFk} & 99.151 & 99.998 & 92.767 & 0.020 & 0.577 \\
SalUn\citep{salun} & 92.189 & 98.539 & 91.860 & 0.287 & 0.555 \\
SCRUB\citep{SCRUB_NegGrad+} & 99.135 & 99.407 & 92.790 & 0.014 & 0.561 \\
BT\citep{bad_teacher} & 91.703 & 99.287 & 90.300 & 0.633 & 0.608 \\
$l1$-sparse\citep{l1sparse/MIAp} & 92.098 & 98.020 & 91.165 & 0.140 & 0.548 \\
MUNBa \citep{munba}& 94.312 & 99.998 & 94.376 & 0.290 & 0.593 \\
DELETE \citep{delete} & 97.861 & 98.189 & 88.568 & 0.061 & 0.578 \\
\bottomrule
\end{tabular}%
}
\label{tab:cifar10-ele10-perf}
\vspace{-2mm}
\end{table*}

\begin{table*}[t]
\caption{Unlearning performance on Half Class 30\% unlearning scenario. $\mathcal{D}_r^e$ and $\mathcal{D}_{test}^f$ represents the entangled retain set and test set with class labels 0,1 and 2. $\mathcal{D}_r^r$ and $\mathcal{D}_{test}^r$ are the same but with labels 3-9. }
\centering
\resizebox{0.49\linewidth}{!}{%
\begin{tabular}{ccccccc}
\toprule
CIFAR10 & $\mathcal{D}_f (\downarrow)$ & $\mathcal{D}_r^e(\uparrow)$ & $\mathcal{D}_r^r(\uparrow)$ & $\mathcal{D}_{test}^f(\uparrow)$ & $\mathcal{D}_{test}^r(\uparrow)$ & $\mathbf{MIA}^e(\uparrow)$ \\
\midrule
Pretrained   (Original) & 99.600 & 99.319 & 99.479 & 95.633 & 94.114 & 0.008 \\
Retrain & 82.548 & 100.000 & 100.000 & 82.000 & 89.200 & 0.561 \\
\textbf{OPC}   (ours) & \textbf{0.667} & \textbf{99.482} & 100.000 & 73.400 & 94.014 & 1.000 \\
\midrule
GA   \cite{GA} & 0.533 & 0.548 & 86.727 & 0.467 & 83.329 & 0.991 \\
RL   \cite{RL} & 91.585 & 99.882 & 99.886 & 82.333 & 88.586 & 0.238 \\
BE   \cite{BEBS} & 29.333 & 28.948 & 95.489 & 25.133 & 90.343 & 0.978 \\
FT   \cite{FT} & 81.911 & 90.489 & 97.197 & 78.367 & 90.300 & 0.297 \\
NGD   \cite{NGD} & 82.919 & 95.363 & 98.543 & 80.133 & 89.929 & 0.285 \\
NegGrad+   \cite{SCRUB_NegGrad+} & 6.919 & 12.444 & 99.651 & 6.867 & 93.071 & 0.980 \\
EUk   \cite{EUk/CFk} & 91.630 & 93.763 & 99.508 & 86.367 & 92.143 & 0.133 \\
CFk   \cite{EUk/CFk} & 96.178 & 99.496 & 99.991 & 90.333 & 93.086 & 0.063 \\
SalUn   \cite{salun} & 86.444 & 98.904 & 99.108 & 81.367 & 85.571 & 0.326 \\
SCRUB   \cite{SCRUB_NegGrad+} & 58.237 & 58.844 & 98.152 & 53.767 & 92.386 & 1.000 \\
BT   \cite{bad_teacher} & 16.667 & 99.274 & 99.476 & 82.933 & 94.471 & 1.000 \\
$l1$-sparse   \cite{l1sparse/MIAp} & 78.193 & 89.867 & 97.175 & 75.033 & 88.486 & 0.367 \\
MUNBa   \cite{munba} & 88.163 & 91.659 & 94.378 & 83.900 & 86.643 & 0.226 \\
DELETE   \cite{delete} & 0.000 & 0.296 & 99.416 & 0.900 & 95.000 & 1.000 \\ 
\bottomrule
\end{tabular}%
}
\resizebox{0.48\linewidth}{!}{%
\begin{tabular}{ccccccc}
\toprule
SVHN & $\mathcal{D}_f (\downarrow)$ & $\mathcal{D}_r^e(\uparrow)$ & $\mathcal{D}_r^r(\uparrow)$ & $\mathcal{D}_{test}^f(\uparrow)$ & $\mathcal{D}_{test}^r(\uparrow)$ & $\mathbf{MIA}^e(\uparrow)$ \\
\midrule
Pretrained   (Original) & 99.531 & 99.569 & 99.157 & 95.142 & 90.984 & 0.008 \\
Retrain & 87.965 & 99.735 & 99.868 & 86.954 & 85.299 & 0.204 \\
\textbf{OPC}   (ours) & \textbf{3.750} & \textbf{99.486} & 99.927 & 74.145 & 92.354 & 1.000 \\
\midrule
GA   \cite{GA} & 61.869 & 61.793 & 91.406 & 51.901 & 81.543 & 0.462 \\
RL   \cite{RL} & 91.042 & 93.007 & 94.790 & 89.047 & 90.000 & 0.301 \\
BE   \cite{BEBS} & 0.491 & 0.756 & 93.275 & 0.355 & 75.246 & 0.997 \\
FT   \cite{FT} & 88.668 & 94.519 & 98.534 & 86.008 & 92.420 & 0.240 \\
NGD   \cite{NGD} & 85.349 & 91.669 & 97.493 & 83.352 & 91.589 & 0.283 \\
NegGrad+   \cite{SCRUB_NegGrad+} & 0.491 & 1.315 & 87.194 & 0.528 & 81.084 & 0.991 \\
EUk   \cite{EUk/CFk} & 95.978 & 99.683 & 99.777 & 90.757 & 90.791 & 0.075 \\
CFk   \cite{EUk/CFk} & 95.381 & 97.430 & 96.717 & 90.193 & 87.287 & 0.086 \\
SalUn   \cite{salun} & 96.641 & 99.728 & 99.899 & 95.342 & 94.056 & 0.104 \\
SCRUB   \cite{SCRUB_NegGrad+} & 7.242 & 7.159 & 91.345 & 3.830 & 86.676 & 0.893 \\
BT   \cite{bad_teacher} & 12.746 & 99.509 & 99.121 & 86.663 & 92.214 & 1.000 \\
$l1$-sparse   \cite{l1sparse/MIAp} & 87.043 & 96.122 & 99.068 & 85.762 & 92.035 & 0.213 \\
MUNBa   \cite{munba} & 91.594 & 95.449 & 97.323 & 90.293 & 94.555 & 0.217 \\
DELETE   \cite{delete} & 2.253 & 5.821 & 99.263 & 2.256 & 91.868 & 0.973 \\
\bottomrule
\end{tabular}%
}
\label{tab:cifar10-cls30ele50-perf}
\vspace{-4mm}
\end{table*}

The performance of OPC as an MU method can be found in \cref{tab:cifar10-cls30-perf}, \cref{tab:cifar10-ele10-perf} and \cref{tab:cifar10-cls30ele50-perf}. 
On all forgetting scenarios, OPC shows superior forgetting ability while retaining RA and TA on both CIFAR10 and SVHN.

For random unlearning, OPC shows superior forgetting ability with FA $<10\%$ while retaining RA and TA on both CIFAR10 and SVHN, while all others fail to reduce FA below 90\%.
In half-class unlearning, OPC is the only MU method which achieve both $FA<10\%$ and $>99\%$ on $\mathcal{D}_r^e$. Other methods except OPC and BT, fails to forget $\mathcal{D}_f$ (e.g. retrain, FT) or forget both $\mathcal{D}_f$ and $\mathcal{D}_r^e$. 
The superior forgetting efficacy follows directly from the strong forget-retain separation as shown in \cref{subsec:OPC-tsne}, such that high MU efficacy occurs only when the entanglement is resolved. 

Since OPC exhibits stronger forgetting compared to retraining, the privacy concerns regarding membership inference attacks are expected; thus, we show the $MIA^p$ score (using \cite{l1sparse/MIAp}'s implementation), which measures potential privacy leakage, in \cref{tab:cifar10-ele10-perf}. 
OPC shows a slightly larger $MIA^p$ score (2.4\% higher $MIA^p$ accuracy for SVHN) compared to retraining. However, considering the UA gain (7.1\% for retrain vs 92.5\% for OPC), we consider this tradeoff to be acceptable.

\vspace{-1mm}
\subsection{Resistance against FM-recovery}\label{subsec:OPC-FMR}
\vspace{-1.5mm}

\begin{figure*}[t]
    \includegraphics[width=\linewidth]{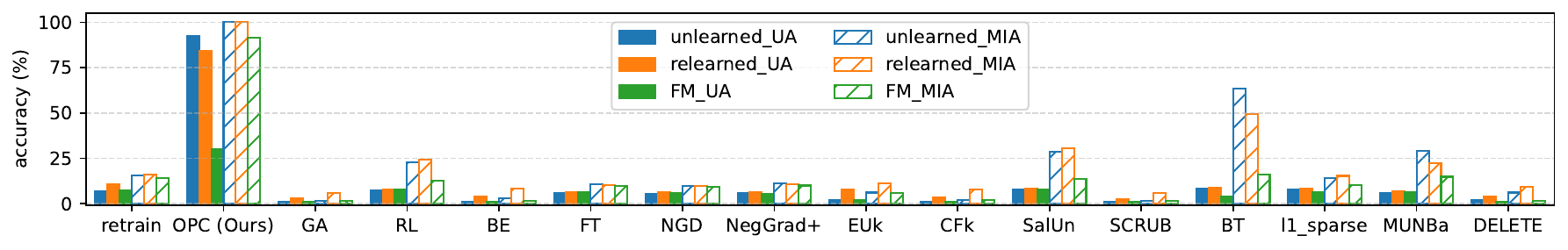}
    \vspace{-0.6cm}
    \caption{Effect of FM recovery in MU performance for SVHN random 10\% unlearning scenario. The blue and orange colored bars represent $UA$ (unlearning accuracy, $100\%-Acc(\mathcal{D}_f)$)  and $MIA^e$, respectively, before and after the FM-recovery.}
    \label{fig:recovery_UA_svhnele10}
    \vspace{-0.4cm}
\end{figure*}

\begin{figure*}[t]
    \includegraphics[width=\linewidth]{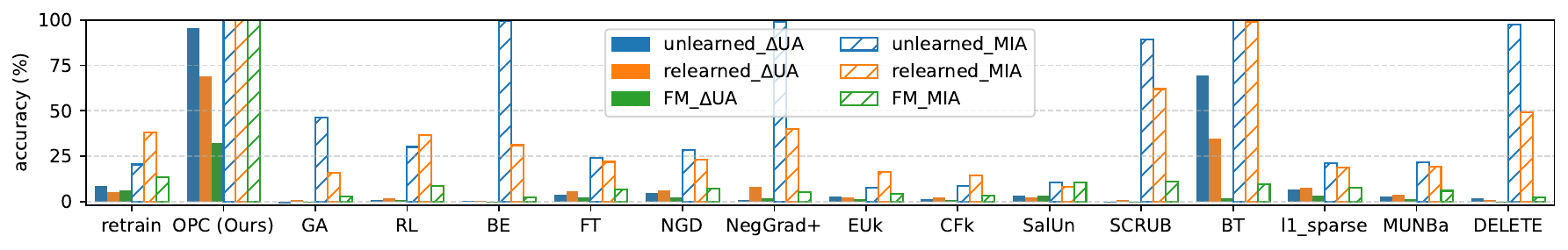}
    \vspace{-0.6cm}
    \caption{Effect of relearning (orange) and \FMR{} (green) compared to unlearned model (blue) for SVHN half-class 30\% unlearning scenario. $\Delta UA$ = $Acc(\mathcal{D}_r^e)-Acc(\mathcal{D}_f)$ represent difference between entangled retain accuracy and forget accuracy, on classess 0,1 and 2.}
    \label{fig:recovery_UA_svhncls30ele50}
    \vspace{-0.4cm}
\end{figure*}

We apply \FMR{} to OPC for evaluation, with results shown in \cref{fig:recovery_UA}, \cref{fig:recovery_UA_svhnele10} and \cref{fig:recovery_UA_svhncls30ele50}; detailed numerical results can be found in \cref{appendix:more_FMR_results}. For \cref{fig:recovery_UA_svhncls30ele50}, we report the difference $\Delta UA$ = $Acc(\mathcal{D}_r^e)-Acc(\mathcal{D}_f)$ since half-class unlearning requires to forget $\mathcal{D}_f$ only, while preserving performance on $\mathcal{D}_r^e$ (entangled retain set). 

While all other MU baselines were vulnerable to FM-recovery, OPC demonstrates superior resistance. In the class unlearning scenario, the post-recovery UA is 55\%, whereas other methods reverted to a UA of less than 15\%. Note that the retrained model shows resistance with a recovered UA of 30\%.

In the random and half-class unlearning scenarios, FM-recovery induces some performance restoration even in OPC; however, OPC still demonstrates the strongest forgetting efficacy. 
Although the UA of OPC drops to 30\%, it remains significantly higher than that of other MU methods, which fail to resolve the forget-retain entanglement, yielding a UA of less than 10\%. 
The BT \cite{bad_teacher} had resolved the forget-retain entanglement and shown promising performance, but due to linear separability of forget features it was easily reverted via \FMR{}. 

In qualitative evaluation via generative decoder, we observe that even if the unlearned features are correctly classified, reconstructing images from OPC-unlearned features is impossible. 
To show this, we sample images from forget-set, that are correctly classified by the FM-recovered OPC model and reported reconstruction results from the unlearned (and recovered) features in \cref{fig:recovery-recon}.

Consequently, even if FM-recovery can restore correct classification predictions, the features unlearned by OPC yield completely different images via the decoder, with almost all details destroyed. Note that this destruction does not occur in retain-set; OPC selectively deletes information  only pertaining to the forget-set.

\vspace{-1mm}
\subsection{Gradient contraction of OPC}\label{subsec:OPC-inversion}
\vspace{-1.5mm}

\begin{figure*}[t]
  \centering
  \begin{subfigure}[t]{0.99\linewidth}
      \includegraphics[width=\textwidth]{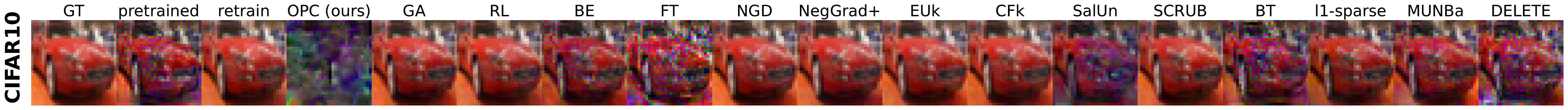}
      \label{subfig:inversion_cifar10cls30}
      \includegraphics[width=\textwidth]{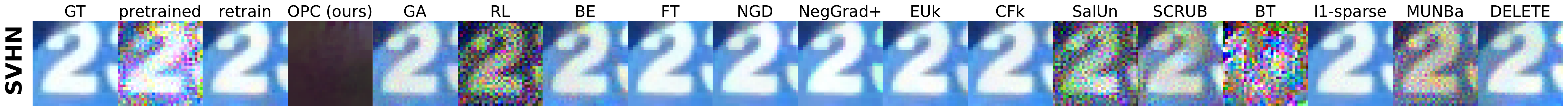}
      \label{subfig:inversion_svhncls30}
    \end{subfigure}
    
\vspace{-0.6cm}
\caption{The reconstruction results of unlearning inversion attack in \cref{subsec:OPC-inversion}. GT represents the ground truth image from the forget set of each dataset and others are the results from each unlearned model.}
\label{fig:inversion-recon}
\vspace{-0.6cm}
\end{figure*}

The OPC's principle, the feature contraction toward origin, invokes collapse of gradient signal. In backpropagation of cross-entropy loss, the gradient of last layer $W$ is given by
 $\nabla_W\mathcal{L}_{CE}=(\hat{y}-y)z^T$, where $y$ is one-hot encoding of class label and $z=f_\theta(x)$ be the feature. 
 Since OPC process pushes $z\rightarrow 0$, the gradient also collapses, rendering all weight gradients uninformative during backpropagation.
This gradient collapse on OPC makes the depth of forgetting deeper: it makes OPC much more resistant to inversion and relearning attack.

\vspace{-1mm} 
\paragraph{Relearning attack on OPC}

In early stages of relearning process, the gradient on unlearned sample would vanish due to contraction, inherently slow down the relearning process. In \cref{fig:recovery_UA}, \cref{fig:recovery_UA_svhnele10} and \cref{fig:recovery_UA_svhncls30ele50}, OPC shows strong resistance on all unlearning scenario.
Especially on CIFAR10 30\% class unlearning, the relearned forget accuracy was 36\%: which is close to random classifier performance on 3 forget classes and support the fatal collapse of feature-level information. 

\vspace{-1mm} 
\paragraph{Unlearning inversion attack on OPC}

Unlearning inversion attack\cite{hu2024_reconattack} aims to reconstruct forgotten input, based on gradient inversion. Since OPC-unlearned model would produce uninformative gradient on forget sample, OPC is naturally resistant to this reconstruction attack, while all models including the original model are vulnerable; In \cref{fig:inversion-recon}, the inversion attack could reconstruct ground truth image from the gradient oracle.

\vspace{-2mm}
\section{Discussion}\label{sec:discussion}
\vspace{-3mm}

While retraining is often considered the golden standard for MU, it fails to achieve high unlearning accuracy (UA) in random forgetting scenario because model generalization maintains performance on the forget-set. 
This creates a gap between theoretical erasure and practical demands, as users expect a verifiable performance drop on forget queries which retraining cannot justify.
We address this by prioritizing maximal forgetting: maximizing UA while preserving utility on retain-set and test set. 
Unlike prior methods that struggle even on simple benchmarks, OPC is the first to successfully decouple forget and retain features (\cref{subsec:OPC-tsne}), which we identify as the fundamental prerequisite for achieving high UA.

FM-recovery exposes a critical vulnerability overlooked by logit-based metrics: the inherent reversibility of shallowly forgotten features. 
This risk extends to transfer learning, where latent information preserved in the unlearned backbone can leak sensitive data during downstream fine-tuning. 
By failing to erase these underlying representations, prior MU methods provide only a superficial privacy guarantee that is easily compromised in practical deployment.

In class unlearning, many methods exploit trivial solutions by merely penalizing classifier weights for forget classes, which inevitably results in shallow forgetting as the underlying features remain informative. 
In contrast, OPC inherently avoids these solutions because they do not minimize $\mathcal{L}_{OPC}$. 
By constraining the gap between forget and retain class logits, OPC forces the predictive probability toward a uniform distribution, while trivial solution penalizes the forget class logits to have large gap between the retain class logits.

Neural networks naturally map uncertain or out-of-distribution (OOD) inputs to small-normed feature regions \cite{dhamija2018reducing,tack2020csi,huang2021importance,park2023understanding}. 
Ideally, an unlearned model should treat forget data as unseen, mimicking this OOD behavior through low-norm representations and high uncertainty. 
OPC formalizes this transition by breaking feature distinguishability; it artificially transforms the forget-set into an OOD-like state but achieves a more rigorous and total collapse of feature information than standard OOD mapping.
This relationship also clarifies the previously unexplained efficacy of the Bad Teacher (BT) method \cite{bad_teacher} as analyzed in \cref{appendix:BT_analysis}.

\vspace{-2mm}
\section{Conclusion}\label{sec:conclusion}
\vspace{-3mm}

Most machine unlearning methods retain what they claim to forget. 
A single linear map fitted in seconds without access to the forgotten data, as we demonstrate with \FMR{}, reverses 14 representative MU methods and recovers both classification performance and pixel-level image content. 
The information was never gone but kept one layer away from the prediction head. 

OPC closes this gap by acting on the latent representation: contracting forget-set latent features to the origin is provably equivalent to maximizing predictive distribution entropy. 
Hence, OPC admits behavioral and feature-level forgetting as a single objective, and at the same time collapses the gradient signal on forget-set data in any type, by selectively contracting the forget feature. 
Moreover, the same mechanism also resolves forget-retain entanglement issue: in random forgetting and half-class forgetting, only OPC succeeds in disentangling forget sets into a low-norm cluster.

More broadly, we note that output-level metrics alone are insufficient to evaluate machine unlearning methods. 
We hope FMR becomes a routine safeguard against shallow forgetting, and that our insights from OPC open a new class of MU algorithms with greater robustness. 

\bibliography{aimlk,refs,refs-dhl}
\bibliographystyle{unsrt}

\appendix
\onecolumn
\newpage
\counterwithin{equation}{section}
\counterwithin{figure}{section}
\counterwithin{table}{section}

\paragraph{\Large Appendices}

\section{experimental details}\label{appendix:exp-details}

\subsection{Experiment settings}\label{appendix:setup-details}

\begin{table*}[h]
    \centering
    \caption{Table of training information on 30\% Class unlearning scenario}
    \resizebox{0.9\linewidth}{!}{%
    \begin{tabular}{ccccccccccccc}
    \toprule
 & \multicolumn{2}{c}{CIFAR10 class 30\%} & \multicolumn{2}{c}{CIFAR10 random 10\%} & \multicolumn{2}{c}{CIFAR10 half-class 30\%} & \multicolumn{2}{c}{SVHN class 30\%} & \multicolumn{2}{c}{SVHN random 10\%} & \multicolumn{2}{c}{SVHN half-class 30\%} \\
 & best epoch & best LR & best epoch & best LR & best epoch & best LR & best epoch & best LR & best epoch & best LR & best epoch & best LR \\
 \midrule
Retrain & 182 & 0.01 & 182 & 0.01 & 0.01 & 182 & 182 & 0.01 & 182 & 0.01 & 0.0002 & 180 \\
OPC (ours) & 30 & 0.01 & 25 & 0.01 & 0.01 & 21 & 20 & 0.009 & 5 & 0.0008 & 0.01 & 28 \\
GA\citep{GA} & 10 & 0.00004 & 5 & 0.000005 & 0.00002 & 30 & 15 & 0.0001 & 15 & 0.0001 & 0.000002 & 30 \\
RL\citep{RL} & 15 & 0.018 & 15 & 0.013 & 0.07 & 30 & 20 & 0.008 & 15 & 0.013 & 0.005 & 1 \\
BE\citep{BEBS} & 10 & 0.0001 & 4 & 0.0000185 & 0.00002 & 30 & 8 & 0.00001 & 4 & 0.000008 & 0.000008 & 30 \\
FT\citep{FT} & 20 & 0.035 & 20 & 0.035 & 0.07 & 3 & 40 & 0.1 & 40 & 0.1 & 0.035 & 3 \\
NGD\citep{NGD} & 20 & 0.035 & 20 & 0.035 & 0.1 & 5 & 40 & 0.1 & 40 & 0.1 & 0.035 & 3 \\
NegGrad+\citep{SCRUB_NegGrad+} & 20 & 0.035 & 15 & 0.035 & 0.035 & 27 & 40 & 0.05 & 10 & 0.03 & 0.04 & 30 \\
EUk\citep{EUk/CFk} & 20 & 0.035 & 20 & 0.035 & 9 & 30 & 40 & 0.1 & 10 & 0.03 & 0.2 & 8 \\
CFk\citep{EUk/CFk} & 20 & 0.04 & 40 & 0.1 & 0.5 & 22 & 40 & 0.1 & 10 & 0.03 & 0.2 & 2 \\
SalUn\citep{salun} & 20 & 0.02 & 15 & 0.015 & 0.1(mask0.1) & 20 & 20 & 0.01 & 15 & 0.01 & 0.177(mask0.5) & 30 \\
SCRUB\citep{SCRUB_NegGrad+} & 3 & 0.0003 & 15 & 0.00007 & 0.0005 & 1 & 3 & 0.002 & 5 & 0.000038 & 0.00005 & 2 \\
BT\citep{bad_teacher} & 5 & 0.01 & 8 & 0.01 & 0.01 & 28 & 12 & 0.01 & 2 & 0.005 & 0.01 & 30 \\
$l1$-sparse\citep{l1sparse/MIAp} & 20 & 0.005 & 20 & 0.015 & 0.01 & 18 & 25 & 0.01 & 20 & 0.01 & 0.01 & 27 \\
MUNBa\citep{munba} & 10 & 0.012 & 10 & 0.015 & 0.05 & 2 & 10 & 0.05 & 10 & 0.05 & 0.2 & 4 \\
DELETE\citep{delete} & 20 & 0.001 & 20 & 0.001 & 0.01 & 30 & 20 & 0.00001 & 10 & 0.000002 & 0.00005 & 30 \\
\bottomrule
\end{tabular}%
}
    \label{tab:unlearn_hparam_class30_lr_epochs}
\end{table*}

In this section, we detail the experimental settings. All experiments were conducted on a machine equipped with an AMD Ryzen 9 5900X 12-Core CPU, an NVIDIA GeForce RTX 3090 GPU with 24GB of VRAM, and 64GB of TEAMGROUP UD4-3200 RAM (2 × 32GB). 
To obtain the pretrained models, we trained ResNet-18 \citep{resnet} from scratch on CIFAR-10 \citep{krizhevsky2010cifar} and SVHN \citep{svhn} datasets. 
The pretrained model was trained for 182 epochs with a learning rate of 0.1 on CIFAR-10, and for 200 epochs with a learning rate of 0.1 on SVHN. The optimizer used in our experiments was Stochastic Gradient Descent (SGD) with a momentum of 0.9 and a weight decay of 1e-5. For learning rate scheduling, we employed PyTorch's MultiStepLR with milestones set at epochs 91 and 136, and a gamma value of 0.1.

For data augmentation, we applied common settings consist of RandomCrop(32, 4) and RandomHorizontalFlip, from the torchvision \citep{torchvision2016} library to CIFAR-10 \citep{torchvision2016}. No augmentation was used for SVHN, considering its digit-centric nature and the presence of multiple digits in a single image, with only the center digit serving as the target. 
Unless otherwise stated, we used a batch size of 256 for all unlearning procedures, including pretraining and retraining. 
However, in DELETE\cite{delete}, we trained with a batch size of 128 as described in their paper.

Based on these settings, we trained unlearned models from pretrained model, with maximal training budget 40 epoch. Since almost all models were showing catastrophic forgetting with severe degradation on RA, we saved intermediate checkpoints and choose best model.
We tuned learning rate until the model shows high UA without degradation of RA and TA. The resulting learning rates and best epoch for each MU are summarized in \cref{tab:unlearn_hparam_class30_lr_epochs}.

\begin{table}[h]
\caption{Table of hyperparameters on unlearning scenario}
\centering
\resizebox{0.99\linewidth}{!}{%
\begin{tabular}{cccccc}
\toprule
Methods &
  \textbf{Hparam name} &
  \textbf{Description of hyperparameters} &
  \textbf{30\% Class} &
  \textbf{10\% random} & \textbf{30\% halfclass} \\ \midrule
\multirow{2}{*}{OPC(Ours)}         & $\lambda_r$ & weight for the cross-entropy loss on retain data,                  & 1                 & 0.95    &1               \\
                                   & $\lambda_f$ & weight for the norm loss on forget data                            & 0.7               & CIFAR10:0.05, SVHN:0.2 & CIFAR10:0.1, SVHN:0.3\\\midrule
NGD\citep{NGD} &
  $\sigma$ &
  standard deviation of Gaussian noise added to gradients &
  $10^{-7}$ &
  $10^{-7}$ &$10^{-7}$\\\midrule
NegGrad+\citep{SCRUB_NegGrad+} &
  $\alpha$ &
  controls weighted mean of retain and forget losses &
  0.999 &
  0.999 &0.999\\\midrule
EUk\citep{EUk/CFk} & $k$         & Last $k$ layers to be trained                                      & 3                 & 3             &3         \\\midrule
CFk\citep{EUk/CFk} & $k$         & Last $k$ layers to be trained                                      & 3                 & 3              &3        \\\midrule
SalUn\citep{salun} & $pt$        & sparsity ratio for weight saliency                                 & 0.5               & 0.5              &CIFAR10:0.1,SVHN:0.5      \\\midrule
\multirow{5}{*}{SCRUB\citep{SCRUB_NegGrad+}} &
  $\alpha$ &
  weight of KL loss between student and teacher. &
  0.001 &
  0.001&0.001 \\
                                   & $\beta$     & scales optional extra distillation loss                            & 0                 & 0         &0             \\
                                   & $\gamma$    & weight of classification loss.                                     & 0.99              & 0.99      &0.99             \\
                                   & $kd\_T$     & controls the softening of softmax outputs for distillation.        & 4                 & 4       &4               \\
                                   & $msteps$    & $\#$ of maximize steps using forget data before minimize training. & CIFAR10:2, SVHN:1 & 1    &1                  \\\midrule
$l1$-sparse\citep{l1sparse/MIAp} &
  $\alpha$ &
  weight of $l1$ regularization &
  0.0001 &
  0.0001 &0.0001 \\\midrule
  MUNBa\citep{munba} &
  $\alpha$ &
  weight of forget loss after bargaining &
  0.1 &
  0.1 &0.1 \\
  \bottomrule
\end{tabular}\label{tab:unlearn_hparam_explanation}
}
\end{table}

Other hyperparameters and their descriptions are provided in \cref{tab:unlearn_hparam_explanation}.

\subsection{MU baselines in benchmark}\label{appendix:methods}

Gradient Ascent (GA) attempts to undo learning from retain set by reversing gradient directions \cite{GA}. Random Labeling (RL) trains the model using retain set and randomly labeled forget set \cite{RL}. Boundary Expanding (BE) pushes forget set to an extra shadow class \cite{BEBS}. Fine Tuning (FT) continues training on retain set using standard stochastic gradient descent (SGD) \cite{FT}. Noisy Gradient Descent (NGD) modifies FT by adding Gaussian noise to each update step \cite{NGD}. Exact Unlearning the last k layers (EUk) retrains only the last k layers from scratch to remove forget set information. Catastrophically Forgetting the last k layers (CFk), instead of retraining, continues training the last k layers on retain set \cite{EUk/CFk}. Saliency Unlearning (SalUn) enhances RL by freezing important model weights using gradient-based saliency maps \cite{salun}. Bad-Teacher (BT) uses a student-teacher framework where the teacher is trained on full train set and the student mimics it for retain set, while imitating a randomly initialized model, the ``bad teacher'', for forget set \cite{bad_teacher}. SCalable Remembering and Unlearning unBound (SCRUB), a state-of-the-art technique, also employs a student-teacher setup to facilitate unlearning. NegGrad+ combines GA and FT to fine-tune the model in a way that effectively removes forget set information \cite{SCRUB_NegGrad+}. $l1$-sparse enhances FT with $l1$ regularization term \cite{l1sparse/MIAp}. Selective Synaptic Dampening (SSD) unlearns by dampening weights that strongly influence the Fisher information of the forget set more than the rest of the dataset \cite{SSD}.

\section{Theory of OPC: remarks and proof}\label{appendix:theory}

The minimization of logit norm $\mathbf{m}_\theta(x)$ is strongly related to the shannon entropy and uncertainty. This connection had been observed in OOD detection literature, that OOD samples exploit high uncertainty with small feature norms.

Ideally, unlearned data should be treated as unseen (OOD) samples, leading the model to exhibit high uncertainty with small feature norms. We formalize this relationship in the following theorem, establishing a lower bound on the predictive entropy as a function of feature norm.

Instead of relying on empirical observations, we formalize this relationship in the following theorem,\cref{thm:entropyLB}, establishing a lower bound on the predictive entropy as a function of feature norm. 
Intuitively, the reason of entropy explosion near origin is due to symmetry: the softmax function on origin returns the uniform distribution and therefore feature close to origin would induce uncertain prediction.

\begin{restatable}{theorem}{ThmentropyLB}\label{thm:entropyLB}
Let $C$ be number of classes. Suppose $\mathbf{h}=\mathbf{m}_\theta(x)\in B_r(0)$ where $B_r(0)$ is the ball of radius $r$ centered at origin. Then the entropy $H(softmax(\mathbf{h}))$ of predicted probability has following lower bound parameterized by $r$ and $C$:

\begin{equation}\label{eq:Hstar}
\begin{aligned}
    &H^*(r,C):=\underset{\mathbf{h}\in B_r(0)}{min} H(softmax(\mathbf{h})) > \log \left(1+(C-1) \exp \left(-\sqrt{\frac{C}{C-1}}r\right)\right)
\end{aligned}
\end{equation}
\end{restatable}

As the feature norm $r$ decreases, the exponential term $\exp\left(-\sqrt{\frac{C}{C-1}}r\right)$ approaches 1, pushing the lower bound in \cref{eq:Hstar} toward $\log(C)$, the maximum possible entropy. 
Conversely, as $r$ increases, the lower bound decreases, reflecting that more confident predictions become available. 

\subsection{proof of \cref{thm:entropyLB}}\label{appendix:entropyLB}

\ThmentropyLB*

   \begin{proof}
        For the clarity, we denote $\mathbf{q}=\exp(\mathbf{h})$ and $\mathbf{y}=softmax(\mathbf{h})=\frac{\mathbf{q}}{\|\mathbf{q}\|_1}$.
   
        Let $X=\exp(B_r(0))$ in $\mathbf{q}$-space and $Y=softmax(B_r(0))$ in $\mathbf{y}$-space.
        Since entropy function $H$ is concave in $\mathbf{y}$-space, the minimal solution $\mathbf{y}^*=argmin H(\mathbf{y})$ must lie in the boundary of $Y$, $\partial Y$. 

        Since $Y$ is a image of $X$ under projection $\mathbf{q}\mapsto \frac{\mathbf{q}}{\|\mathbf{q}\|_1}$ and thus $H(\frac{\mathbf{q}}{\|\mathbf{q}\|_1})= H(\frac{c\mathbf{q}}{\|c\mathbf{q}\|_1})$ for all $c>0$, the condition $\mathbf{y}^*=\frac{\mathbf{q}^*}{\|\mathbf{q}^*\|_1}\in \partial Y$ would be translated to followings in $\mathbf{q}$-space:
        \begin{enumerate}
            \item $\mathbf{q}^*\in \partial X$
            \item The tangent space $T_{\mathbf{q}^*}X$ includes the origin, $0$.
        \end{enumerate}

    Since $X=\exp(B_r(0))$, the boundary $\partial X$ would be given by
    \begin{equation}
        \partial X = \{\mathbf{q}| \sum_{i=1}^C (\log q_i)^2 =r^2\}
    \end{equation}
    and $T_{\mathbf{q}^*}(X)$ would be 
    \begin{equation}
        T_{\mathbf{q}^*}(X) = \{\mathbf{q}|\sum_{i=1}^C \frac{\log q_i^*}{q_i^*}(q_i-q_i^*)=0\}.
    \end{equation}
    Hence, we get $\sum_{i=1}^C \log q_i^*=0$ since $0\in T_{\mathbf{q}^*}X$. 

    Therefore, we can find $q^*$ by solving the following constrianed optimization problem.
    \begin{equation}
        \begin{aligned}
            \mbox{minimize } &H(\frac{\mathbf{q}}{\|\mathbf{q}\|_1}) \\
            \mbox{subject to }& \sum_{i=1}^C \log q_i =0\\
            & \sum_{i=1}^C (\log q_i)^2 = r^2
        \end{aligned}
    \end{equation}

    Or equlvalently in $\mathbf{h}$-space:
    \begin{equation}
        \begin{aligned}
            \mbox{minimize } &H(softmax(\mathbf{h})) \\
            \mbox{subject to }& \sum_{i=1}^C h_i =0\\
            & \sum_{i=1}^C h_i^2 = r^2
        \end{aligned}.
    \end{equation}

    Since entropy function is Schur-concave\cite{marshall1979inequalities} and Schur-concave functions are minimized at the extreme points of a constraint's majorization lattice. Similar to \cite{brazitikos2025sharp} which studies Schur-concave function on same constraint, the minimizer would be determined to be vectors with exactly $C-1$ equal coordinates and one opposite coordinate.
    Specifically, minimizing Shannon entropy subject to $\sum h_i^2 = r^2$ and $\sum h_i = 0$, the "two-value" solution ($b$ elements take value $h_1$ and $C-b$ elements take value $h_C$ for some $b<C$) is the standard result for extremizing Schur-concave functions under these constraints.

    Now, we can find $h_1$ and $h_C$ from $g_1(\mathbf{h})=g_2(\mathbf{h})=0$ for each $b$ that
    \begin{equation}
        h_1 = \sqrt{\frac{C-b}{bC}}r, h_C=-\sqrt{\frac{b}{C(C-b)}}r
    \end{equation}
    , which are the stationary points of Lagrangian.

    Considering the characteristic of entropy, which is minimized when only one entry is large and rest are small, the optimal $b$ would be $b=1$. 
    This gives the minimizer 
    \begin{equation}
        \mathbf{h}^*=(\sqrt{\frac{C-1}{C}r},-\frac{r}{\sqrt{C(C-1)}},\cdots -\frac{r}{\sqrt{C(C-1)}}).
    \end{equation}

    Letting $u=-\frac{r}{\sqrt{C(C-1)}}$ and $v=\sqrt{\frac{C}{C-1}}r$, we can rewrite $\mathbf{h}^*=(u+v,u,\cdots,u)$ and obtain
    \begin{equation}
        \mathbf{y}^*=(\frac{e^v}{e^v+C-1},\frac{1}{e^v+C-1},\cdots,\frac{1}{e^v+C-1}).
    \end{equation}
    Letting $\kappa=\frac{e^v}{C-1}$, the minimal entropy $H(\mathbf{y}^*)$ is given by
    \begin{equation}
        \begin{aligned}
            H(\mathbf{y}^*) &= -\frac{e^v}{e^v+C-1} (v-\log(e^v+C-1))+(C-1)\frac{\log(e^v+C-1)}{e^v+C-1}\\
            &=\log(e^v+C-1)-\frac{e^vv}{e^v+C-1}\\
            &=\log((\kappa+1)(C-1))-\frac{\kappa(C-1)\log(\kappa(C-1))}{(\kappa+1)(C-1)}\\
            &=\log(\kappa+1)+\log(C-1)-\frac{\kappa}{\kappa+1}(\log(\kappa)+\log(C-1))\\
            &=\frac{\log(C-1)}{\kappa+1}+\log(\frac{\kappa+1}{\kappa})+\frac{\log(\kappa)}{\kappa+1}\\
            &= \log(1+\frac{1}{\kappa})+\frac{\log(\kappa(C-1))}{\kappa+1}.
        \end{aligned}
    \end{equation}

    Since $\kappa>0$ and $\log(\kappa(C-1))=\log(e^v)=\sqrt{\frac{C}{C-1}}r>0$, we have
    \begin{equation}
        H(\mathbf{y}^*)>\log(1+\frac{1}{\kappa})=\log(1+(C-1)e^{-v})=\log(1+(C-1)\exp(-\sqrt{\frac{C}{C-1}}r)).
    \end{equation}

\end{proof}

\subsection{Minimizing feature norm vs logit norm}

Although we discuss the feature separability, our OPC implementation minimizes logit norm $Wz=\mathbf{m}_\theta(x)$ instead of feature norm $z=f_\theta(x)$. The main reason is fair comparison, since the benchmark methods are designing loss function on logit-level and directly manipulating features would not be fair for scientific research.

While not strictly equivalent algebraically, the minimization of the feature norm $\|z\|_2$ and the logit norm $\|Wz\|_2$ are dynamically coupled during optimization. In the forward direction, minimizing $z$ naturally bounds the logits, governed by the spectral norm inequality $\|Wz\|_2 \le \|W\|_2\|z\|_2$. Conversely, actively minimizing $\|Wz\|_2$ reliably drives the global feature norm $\|z\|_2 \to 0$ due to the gradient dynamics over the dataset. The gradient with respect to the features, $\nabla_z\|Wz\|_2 = \frac{1}{\|Wz\|_2}W^TWz$, strictly penalizes any component of $z$ that lies within the row space of $W$. 
Although this gradient mathematically vanishes when $z \in \text{Ker}(W)$ (i.e., $Wz=0$), $z$ does not stall in this null space in practice. 
Because the network is jointly optimized using the retain loss $\mathbb{E}_{x,y\sim\mathcal{D}_r} \mathcal{L}(\mathbf{m}_\theta(x),y)$, the weight matrix $W$ is prevented from collapsing toward zero. 
Furthermore, the continuous updates to $W$ driven by the retain samples cause the null space of $W$ to constantly rotate during training. This rotation continuously exposes all latent dimensions of $z$ to the row space, where they are systematically penalized by the gradient, ultimately forcing the global convergence of $\|z\|_2 \to 0$.

\section{CKA analysis}

\begin{figure*}[h]
    \centering
    \begin{subfigure}[t]{0.48\textwidth}
        \includegraphics[width=\textwidth]{plots/plots_CKA/cifar10_class30/issue27_CKA_pretrained_FMrecovery.pdf}
        \caption{CKA analysis on CIFAR10}
        \label{subfig:CKA_appendix_CIFAR10cls30}
    \end{subfigure}
    \begin{subfigure}[t]{0.48\textwidth}
        \includegraphics[width=\textwidth]{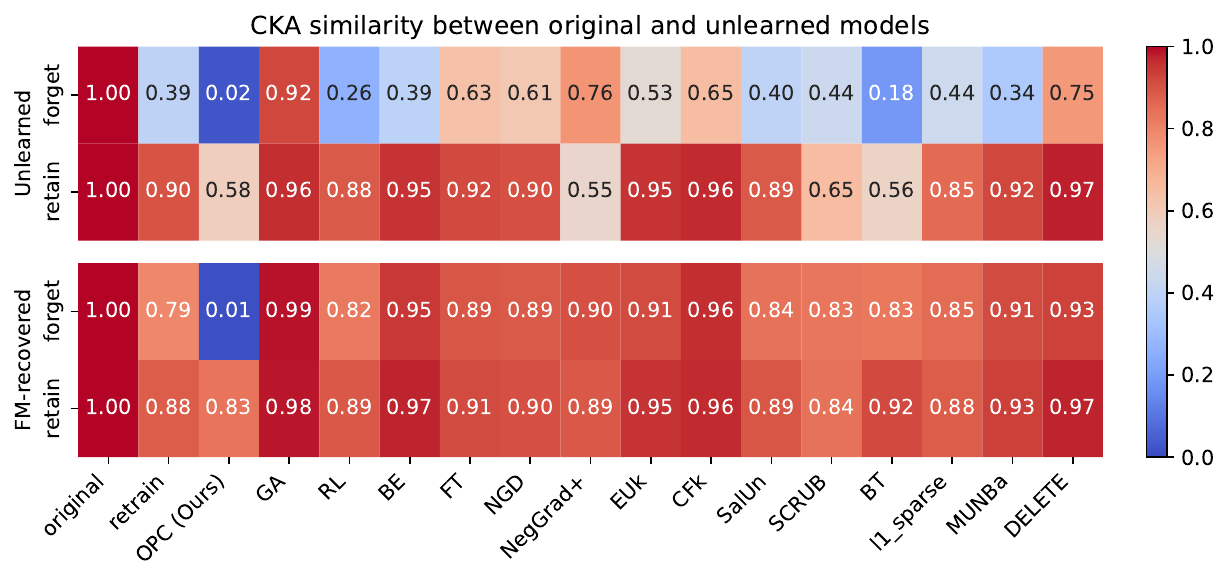}
        \caption{CKA analysis on SVHN}
        \label{subfig:CKA_appendix_SVHNcls30}
    \end{subfigure}
    \caption{CKA similarity between original features and unlearned features in class unlearning scenario, measured on forget-set $\mathcal{D}_f$ and retain-set $\mathcal{D}_r$, before and after FM-recovery.}
\label{fig:appendix-CKA-class}
\end{figure*}

\begin{figure*}[h]
    \centering
    \begin{subfigure}[t]{0.48\textwidth}
        \includegraphics[width=\textwidth]{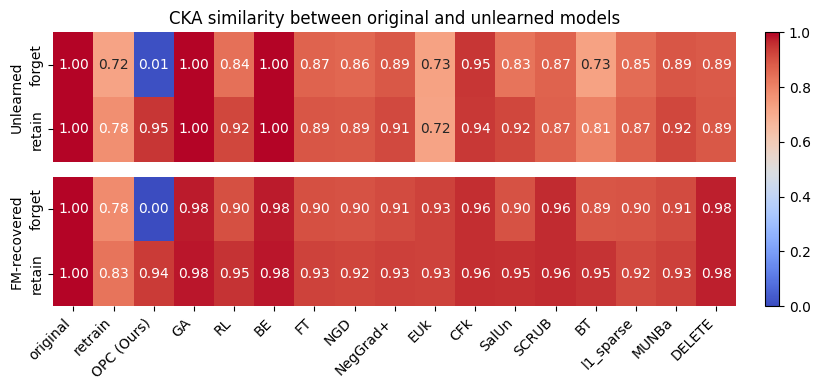}
        \caption{CKA analysis on CIFAR10}
        \label{subfig:CKA_appendix_CIFAR10ele10}
    \end{subfigure}
    \begin{subfigure}[t]{0.48\textwidth}
        \includegraphics[width=\textwidth]{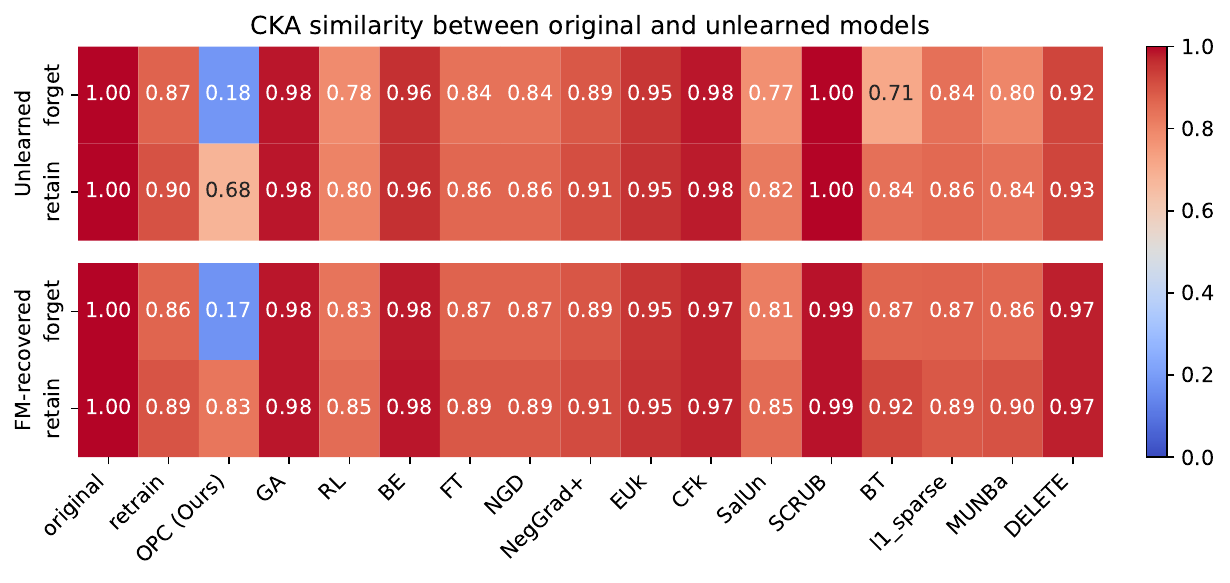}
        \caption{CKA analysis on SVHN}
        \label{subfig:CKA_appendix_SVHNele10}
    \end{subfigure}
    \caption{CKA similarity between original features and unlearned features in random unlearning scenario, measured on forget-set $\mathcal{D}_f$ and retain-set $\mathcal{D}_r$, before and after FM-recovery.}
\label{fig:appendix-CKA-element}
\end{figure*}

As \cite{kim2026we} exploited, the CKA similarity between unlearned features and original or retrained feature on forget-set can evaluate the forgetting efficacy in representation level. However, as discussed in \cref{subsec:recovery-CKA} the FM-recovery makes CKA similarity gets closer to 1, indicating the CKA cannot detect the shallowness in perspective of reversibility. In \cref{fig:appendix-CKA-class} and \cref{fig:appendix-CKA-element} we show the full CKA analysis results.

Consistent to \cref{fig:recovery-cifar10cls30-CKA}, the FM-recovery shows remarkable increase of CKA similarity in class unlearning scenario, that the recovered features are close to original features, while OPC shows strong resistance. For random unlearning scenario, the increase of CKA by FM-recovery was observed with smaller difference, but it is natural regarding the low UA gain of each MU method.


\section{More experimental results}\label{appendix:more-experiments}
In this section, we list additional results omitted from main article due to page limit. We also include several results on larger benchmark: unlearning ViT-B-32 on TinyImagenet dataset.

\subsection{tSNE visualization}\label{appendix-subsec:more_tSNE}

We show tSNE visualization on SVHN 30\% class unlearning and CIFAR10 10\% random unlearning. The CIFAR10 30\% class unlearning result and SVHN 10\% result can be found in \cref{fig:tsne-feature-class-cifar10} and \cref{fig:tsne-feature-rand-svhn}.

\begin{figure*}[h]
\centering
    \includegraphics[width=0.98\textwidth]{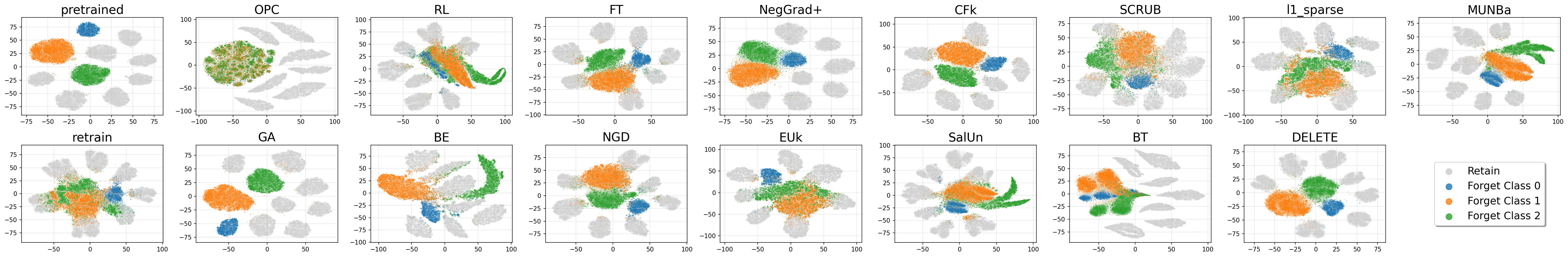}
    \caption{tSNE visualization of unlearned features on random 10\% unlearning scenario of SVHN experiment. The gray-colored points represent retain features and forget features were colored along its class labels.}
    \label{fig:tsne-feature-cls-svhn}
\end{figure*}

\begin{figure*}[h]
\centering
    \includegraphics[width=0.98\textwidth]{plots/plots_tSNE/tsne_comparison_cifar10ele10_feature.png}
    \caption{tSNE visualization of unlearned features on random 10\% unlearning scenario of CIFAR10 experiment. The gray-colored points represent retain features and forget features were colored along its class labels.}
    \label{fig:tsne-feature-rand-cifar}
\end{figure*}

\begin{figure*}[h]
\centering
    \includegraphics[width=0.98\textwidth]{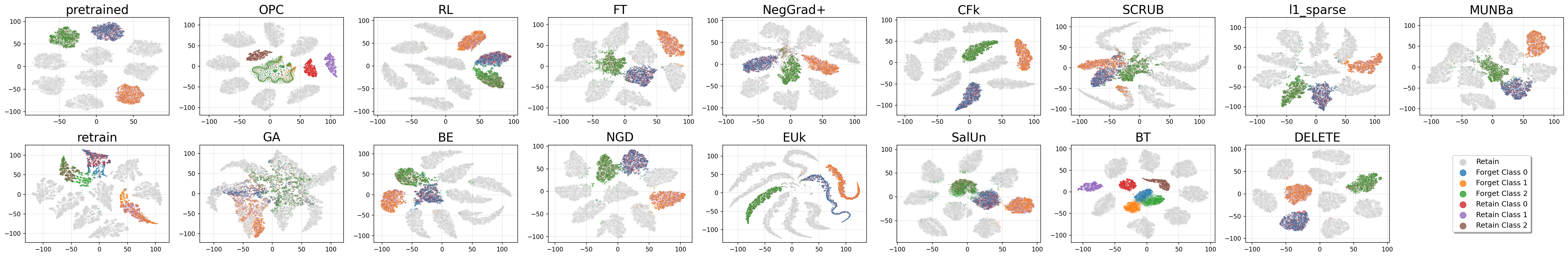}
    \caption{tSNE visualization of unlearned features on half-class 30\% unlearning scenario of CIFAR10 experiment. The gray-colored points represent retain features with class 3-9, and forget features and retain features in class 0-2 were colored along its class labels.}
    \label{fig:tsne-feature-halfclass-cifar}
\end{figure*}

For the SVHN class unlearning results in \cref{fig:tsne-feature-cls-svhn}, the results are consistent to CIFAR10: OPC and retraining are mixing forget features to prevent separability, while other MU are showing clusters along true label and separability.

The OPC behavior on CIFAR10 random and half-class unlearning scenario consistent to SVHN, that OPC the forget features are clearly separated from retain cluster, and linearly inseparable. By resolving the data entanglement, OPC uniquely achieves selective forgetting without catastrophic forgetting while others including retraining failed.

In half-class unlearning scenario,  BT\cite{bad_teacher} succeeded to detach the forget features from entangled retain cluster and achieved good performance (16.67\% FA while retaining accuracy on $\mathcal{D}_r^e$>99\%). However, since the forget features have linear separability along their true labels, \FMR{} could perfectly undo this unlearning: FA was recovered to 94.81\% while retaining $Acc(\mathcal{D}_r^e)>99\%$.

\subsection{MU performance}
The MU performances for CIFAR10 and SVHN experiments can be found in \cref{tab:cifar10-cls30-perf} and \cref{tab:cifar10-ele10-perf}. In this section, we show MU performance of TinyImagenet results in \cref{tab:tinyimagenet-cls10-perf}.

\begin{table}[h]

\caption{Unlearning performance on TinyImageNet}
\centering
\resizebox{!}{1.3cm}{%
\begin{tabular}{cccccc}
\toprule
\textbf{Class 10\%} & Train $\mathcal{D}_f(\downarrow)$ & Train $\mathcal{D}_r(\uparrow)$ & test $\mathcal{D}_f(\downarrow)$ & test $\mathcal{D}_r(\uparrow)$ & $\mathbf{MIA}^e(\uparrow)$ \\
\midrule
Pretrained & 97.830 & 97.541 & 85.200 & 83.685 & 0.105 \\
Retrain & 0.000 & 95.844 & 0.000 & 82.818 & 1.000 \\
OPC (ours) & 0.660 & 99.427 & 0.400 & 81.129 & 1.000 \\
\midrule
RL\citep{RL} & 3.690 & 99.953 & 2.200 & 81.974 & 1.000 \\
FT\citep{FT} & 16.490 & 99.977 & 14.600 & 80.596 & 1.000 \\
SSD\citep{SSD} & 4.730 & 95.800 & 4.800 & 82.263 & 1.000 \\
SalUn\citep{salun} & 3.240 & 99.941 & 2.000 & 82.040 & 1.000 \\
\bottomrule
\end{tabular}%
}
\resizebox{!}{1.3cm}{%
\begin{tabular}{cccccc}
\toprule
\textbf{Random 10\%} & $\mathcal{D}_f (\downarrow)$ & $\mathcal{D}_r(\uparrow)$ & $\mathcal{D}_{test}(\uparrow)$ & $\mathbf{MIA}^e(\uparrow)$ &$\mathbf{MIA}^p(\downarrow)$ \\
\midrule
Pretrained & 97.520 & 97.576 & 83.837  & 0.119 & 0.604 \\
Retrain & 85.930 & 98.682 & 85.337 & 0.276 & 0.606 \\
OPC (ours) & 13.750 & 99.711 & 78.0356 & 0.990 & 0.651  \\
\midrule
RL\citep{RL} & 93.330 & 98.803 & 82.376 & 0.422 & 0.631  \\
FT\citep{FT} & 89.590 & 99.944 & 80.836  & 0.240 & 0.663 \\
SSD\citep{SSD} & 97.350 & 97.356 & 83.597 & 0.128 & 0.600  \\
SalUn\citep{salun} & 94.840 & 98.567 & 82.416 & 0.461 & 0.628 \\
\bottomrule
\end{tabular}%
}
\label{tab:tinyimagenet-cls10-perf}
\end{table}

While other standard baselines were unavailable to induce high efficacy in forgetting, OPC shows superior MU performance, on both class unlearning and random unlearning scenarios.

\subsection{FM-recovery performance}\label{appendix:more_FMR_results}
In this section, we show full table of FM-recovery results, whose MU and $MIA^e$ were visualized in \cref{fig:recovery_UA} and \cref{fig:recovery_UA_svhnele10}.

\begin{table*}[h]

\caption{Recovered performance with FM-recovery and original head on 30\% Class unlearning scenario}
\resizebox{0.5\linewidth}{!}{%
\begin{tabular}{cccccc}
\toprule
\textbf{CIFAR10} & Train $\mathcal{D}_f(\downarrow)$ & Train $\mathcal{D}_r(\uparrow)$ & test $\mathcal{D}_f(\downarrow)$ & test $\mathcal{D}_r(\uparrow)$ & $\mathbf{MIA}^e(\uparrow)$ \\
\midrule
Pretrained & 99.444 & 99.416 & 94.800 & 94.400 & 0.015 \\
Retrain & 70.341 & 95.435 & 70.400 & 86.700 & 0.556 \\
OPC (ours) & 45.000 & 99.000 & 44.200 & 90.929 & 0.944 \\
\midrule
GA\citep{GA} & 86.622 & 96.010 & 81.733 & 90.500 & 0.283 \\
RL\citep{RL} & 94.356 & 98.711 & 89.233 & 92.086 & 0.121 \\
BE\citep{BEBS} & 99.400 & 99.413 & 94.533 & 93.857 & 0.022 \\
FT\citep{FT} & 90.644 & 98.390 & 87.800 & 92.186 & 0.235 \\
NGD\citep{NGD} & 89.778 & 98.181 & 85.867 & 92.386 & 0.255 \\
NegGrad+\citep{SCRUB_NegGrad+} & 87.526 & 97.730 & 84.467 & 91.014 & 0.298 \\
EUk\citep{EUk/CFk} & 96.444 & 99.311 & 90.100 & 93.586 & 0.182 \\
CFk\citep{EUk/CFk} & 98.711 & 99.613 & 93.000 & 94.386 & 0.080 \\
SalUn\citep{salun} & 96.081 & 99.432 & 91.333 & 93.314 & 0.092 \\
SCRUB\citep{SCRUB_NegGrad+} & 89.444 & 97.651 & 84.633 & 92.257 & 0.255 \\
BT\citep{bad_teacher} & 99.304 & 99.438 & 93.133 & 94.329 & 0.041 \\
$l1$-sparse\citep{l1sparse/MIAp} & 95.274 & 99.184 & 89.900 & 93.343 & 0.169  \\
MUNBa \citep{munba}& 97.444 & 99.152 & 91.833 & 93.400 & 0.070 \\
DELETE \citep{munba}& 99.348 & 99.422 & 94.367 & 94.229 & 0.029 \\
\bottomrule
\end{tabular}%
}
\resizebox{0.5\linewidth}{!}{%
\begin{tabular}{cccccc}
\toprule
\textbf{SVHN} & Train $\mathcal{D}_f(\downarrow)$ & Train $\mathcal{D}_r(\uparrow)$ & test $\mathcal{D}_f(\downarrow)$ & test $\mathcal{D}_r(\uparrow)$ & $\mathbf{MIA}^e(\uparrow)$ \\
\midrule
Pretrained & 99.531 & 99.172 & 94.960 & 91.110 & 0.009 \\
Retrain & 88.434 & 96.682 & 88.428 & 87.660 & 0.196 \\
OPC (ours) & 51.304 & 99.068 & 50.637 & 90.818 & 1.000 \\
\midrule
GA\citep{GA} & 99.422 & 99.161 & 93.959 & 91.237 & 0.014 \\
RL\citep{RL} & 92.229 & 97.340 & 91.003 & 90.625 & 0.132 \\
BE\citep{BEBS} & 99.369 & 99.073 & 93.313 & 89.535 & 0.024 \\
FT\citep{FT} & 94.769 & 98.278 & 93.777 & 91.150 & 0.100 \\
NGD\citep{NGD} & 94.111 & 97.862 & 93.577 & 91.789 & 0.110 \\
NegGrad+\citep{SCRUB_NegGrad+} & 94.145 & 96.312 & 93.987 & 91.430 & 0.093 \\
EUk\citep{EUk/CFk} & 96.035 & 98.891 & 93.049 & 90.193 & 0.091 \\
CFk\citep{EUk/CFk} & 99.210 & 99.661 & 94.141 & 90.605 & 0.034 \\
SalUn\citep{salun} & 92.482 & 97.292 & 91.257 & 90.658 & 0.125 \\
SCRUB\citep{SCRUB_NegGrad+} & 91.620 & 89.937 & 90.857 & 85.020 & 0.126 \\
BT\citep{bad_teacher} & 94.795 & 98.171 & 92.986 & 89.907 & 0.109 \\
$l1$-sparse\citep{l1sparse/MIAp} & 92.701 & 96.244 & 91.985 & 89.740 & 0.127  \\
MUNBa \citep{munba}& 96.931 & 98.974 & 94.314 & 90.811 & 0.067 \\
DELETE \citep{delete}& 97.747 & 98.817 & 94.078 & 90.426 & 0.067 \\
\bottomrule
\end{tabular}%
\label{tab:cifar10-cls30-featuremap}
}
\end{table*}

\begin{table*}[h]
\caption{Recovered performance with FM-recovery and original head on 10\% random unlearning scenario}
\resizebox{0.5\linewidth}{!}{%
\begin{tabular}{cccccc}
\toprule
\textbf{CIFAR10} & $\mathcal{D}_f (\downarrow)$ & $\mathcal{D}_r(\uparrow)$ & $\mathcal{D}_{test}(\uparrow)$ & $\mathbf{MIA}^e(\uparrow)$ \\
\midrule
Pretrained & 99.356 & 99.432 & 94.520 & 0.015 \\
Retrain & 90.489 & 99.570 & 89.110 & 0.172 \\
OPC (ours) & 10.267&99.965&92.530&1.000 \\
\midrule
GA\citep{GA} & 99.311 & 99.430 & 94.340 & 0.018 \\
RL\citep{RL} & 94.000 & 99.916 & 93.960 & 0.194 \\
BE\citep{BEBS} & 99.333 & 99.437 & 94.380 & 0.016 \\
FT\citep{FT} & 95.511 & 99.728 & 93.200 & 0.114 \\
NGD\citep{NGD} & 96.000 & 99.731 & 93.540 & 0.114 \\
NegGrad+\citep{SCRUB_NegGrad+} & 96.133 & 99.770 & 93.210 & 0.109 \\
EUk\citep{EUk/CFk} & 99.133 & 99.694 & 93.600 & 0.041 \\
CFk\citep{EUk/CFk} & 99.311 & 99.842 & 94.080 & 0.028 \\
SalUn\citep{salun} & 93.889 & 99.896 & 93.810 & 0.200 \\
SCRUB\citep{SCRUB_NegGrad+} & 99.400 & 99.541 & 94.230 & 0.025 \\
BT\citep{bad_teacher} & 93.000 & 99.351 & 93.150 & 0.193 \\
$l1$-sparse\citep{l1sparse/MIAp} & 94.089 & 98.309 & 92.020 & 0.110 \\
MUNBa \citep{munba}& 95.311 & 99.025 & 92.770 & 0.104 \\
DELETE \citep{delete}& 99.244 & 99.400 & 94.070 & 0.020 \\
\bottomrule
\end{tabular}%
}
\resizebox{0.5\linewidth}{!}{%
\begin{tabular}{cccccc}
\toprule
\textbf{SVHN} & $\mathcal{D}_f (\downarrow)$ & $\mathcal{D}_r(\uparrow)$ & $\mathcal{D}_{test}(\uparrow)$ & $\mathbf{MIA}^e(\uparrow)$ \\
\midrule
Pretrained & 99.151 & 99.334 & 92.736 & 0.015 \\
Retrain & 92.826 & 99.978 & 92.390 & 0.141 \\
OPC (ours) & 69.862 & 99.184 & 92.225 & 0.913 \\
\midrule
GA\citep{GA} & 98.878 & 99.316 & 92.498 & 0.016 \\
RL\citep{RL} & 92.356 & 96.153 & 91.772 & 0.125 \\
BE\citep{BEBS} & 99.135 & 99.287 & 92.221 & 0.015 \\
FT\citep{FT} & 93.872 & 99.643 & 94.211 & 0.099 \\
NGD\citep{NGD} & 94.373 & 99.589 & 94.353 & 0.092 \\
NegGrad+\citep{SCRUB_NegGrad+} & 94.449 & 99.916 & 93.977 & 0.100 \\
EUk\citep{EUk/CFk} & 97.952 & 99.975 & 92.425 & 0.059 \\
CFk\citep{EUk/CFk} & 99.151 & 99.993 & 92.836 & 0.022 \\
SalUn\citep{salun} & 92.143 & 97.695 & 91.580 & 0.137 \\
SCRUB\citep{SCRUB_NegGrad+} & 99.151 & 99.388 & 92.717 & 0.014 \\
BT\citep{bad_teacher} & 96.041 & 99.196 & 91.848 & 0.159 \\
$l1$-sparse\citep{l1sparse/MIAp} & 93.781 & 98.910 & 93.147 & 0.103 \\
MUNBa \citep{munba} & 93.645 & 99.946 & 94.080 & 0.148 \\
DELETE \citep{delete} & 99.135 & 99.270 & 91.918 & 0.016 \\
\bottomrule
\end{tabular}%
}
\label{tab:cifar10-ele10-featuremap}
\end{table*}

\begin{table*}[h]
\caption{FM-recovered performance on Half Class 30\% unlearning scenario. $\mathcal{D}_r^e$ and $\mathcal{D}_{test}^f$ represents the entangled retain set and test set with class labels 0,1 and 2. $\mathcal{D}_r^r$ and $\mathcal{D}_{test}^r$ are the same but with labels 3-9. }
\centering
\resizebox{0.49\linewidth}{!}{%
\begin{tabular}{ccccccc}
\toprule
CIFAR10 & $\mathcal{D}_f (\downarrow)$ & $\mathcal{D}_r^e(\uparrow)$ & $\mathcal{D}_r^r(\uparrow)$ & $\mathcal{D}_{test}^f(\uparrow)$ & $\mathcal{D}_{test}^r(\uparrow)$ & $\mathbf{MIA}^e(\uparrow)$ \\
\midrule
Pretrained   (Original)             & 99.600          & 99.319          & 99.479        & 95.633      & 94.114      & 0.008        \\
Retrain                             & 88.341          & 100.000         & 100.000       & 87.500      & 87.314      & 0.283        \\
\textbf{OPC}   (ours)               & \textbf{33.778} & 99.719          & 99.987        & 74.933      & 92.943      & 1.000        \\
\midrule
GA   \cite{GA}                     & 82.978          & 83.807          & 93.686        & 79.633      & 88.171      & 0.330        \\
RL   \cite{RL}                     & 92.385          & 99.956          & 99.876        & 84.533      & 88.129      & 0.199        \\
BE   \cite{BEBS}                   & 97.452          & 96.696          & 97.454        & 91.333      & 91.557      & 0.067        \\
FT   \cite{FT}                     & 92.474          & 98.178          & 98.013        & 89.833      & 90.843      & 0.163        \\
NGD   \cite{NGD}                   & 92.741          & 99.778          & 99.092        & 90.900      & 90.114      & 0.168        \\
NegGrad+   \cite{SCRUB_NegGrad+}   & 95.615          & 99.911          & 99.797        & 92.000      & 92.157      & 0.132        \\
EUk   \cite{EUk/CFk}               & 95.867          & 97.807          & 85.232        & 91.300      & 80.157      & 0.112        \\
CFk   \cite{EUk/CFk}               & 98.533          & 99.956          & 99.946        & 93.433      & 92.771      & 0.052        \\
SalUn   \cite{salun}               & 88.119          & 98.963          & 99.283        & 83.433      & 85.186      & 0.274        \\
SCRUB   \cite{SCRUB_NegGrad+}      & 95.052          & 95.733          & 98.616        & 89.867      & 92.800      & 0.146        \\
BT   \cite{bad_teacher}            & 94.815          & 99.378          & 99.454        & 92.133      & 93.886      & 0.433        \\
$l1$-sparse   \cite{l1sparse/MIAp} & 91.141          & 98.711          & 98.537        & 88.333      & 90.143      & 0.169        \\
MUNBa   \cite{munba}               & 94.667          & 97.407          & 97.035        & 90.667      & 88.943      & 0.114        \\
DELETE   \cite{delete}             & 98.844          & 99.082          & 99.349        & 94.500      & 93.786      & 0.034       \\
\bottomrule
\end{tabular}%
}
\resizebox{0.49\linewidth}{!}{%
\begin{tabular}{ccccccc}
\toprule
SVHN & $\mathcal{D}_f (\downarrow)$ & $\mathcal{D}_r^e(\uparrow)$ & $\mathcal{D}_r^r(\uparrow)$ & $\mathcal{D}_{test}^f(\uparrow)$ & $\mathcal{D}_{test}^r(\uparrow)$ & $\mathbf{MIA}^e(\uparrow)$ \\
\midrule
Pretrained   (Original) & 99.535 & 99.529 & 99.172 & 94.960 & 91.110 & 0.008 \\
Retrain & 91.280 & 97.447 & 98.485 & 90.211 & 84.229 & 0.136 \\
\textbf{OPC}   (ours) & \textbf{59.231} & 91.467 & 99.623 & 82.560 & 90.040 & 1.000 \\
\midrule
GA   \cite{GA} & 99.019 & 99.113 & 99.101 & 93.313 & 90.971 & 0.029 \\
RL   \cite{RL} & 94.952 & 95.583 & 93.670 & 93.213 & 88.564 & 0.087 \\
BE   \cite{BEBS} & 99.353 & 99.372 & 99.076 & 93.259 & 89.395 & 0.022 \\
FT   \cite{FT} & 96.146 & 98.461 & 97.500 & 95.297 & 91.602 & 0.068 \\
NGD   \cite{NGD} & 96.196 & 98.365 & 97.487 & 95.806 & 91.828 & 0.069 \\
NegGrad+   \cite{SCRUB_NegGrad+} & 96.591 & 98.624 & 97.376 & 96.134 & 91.888 & 0.054 \\
EUk   \cite{EUk/CFk} & 98.078 & 99.481 & 99.597 & 93.959 & 90.293 & 0.042 \\
CFk   \cite{EUk/CFk} & 98.887 & 99.493 & 99.367 & 94.642 & 90.625 & 0.030 \\
SalUn   \cite{salun} & 96.763 & 99.535 & 99.329 & 96.197 & 92.693 & 0.051 \\
SCRUB   \cite{SCRUB_NegGrad+} & 93.505 & 93.181 & 94.253 & 90.821 & 88.364 & 0.109 \\
BT   \cite{bad_teacher} & 96.348 & 98.328 & 98.784 & 95.124 & 89.747 & 0.094 \\
$l1$-sparse   \cite{l1sparse/MIAp} & 95.751 & 98.902 & 99.293 & 95.069 & 92.148 & 0.077 \\
MUNBa   \cite{munba} & 96.136 & 97.550 & 96.484 & 96.243 & 93.324 & 0.059 \\
DELETE   \cite{delete} & 99.282 & 99.270 & 99.093 & 93.095 & 90.219 & 0.025 \\ 
\bottomrule
\end{tabular}
}
\label{tab:cifar10-cls30-perf}
\end{table*}

\begin{table*}[h]
\caption{Recovered performance with \FMR{} on TinyImageNet}
\centering
\resizebox{!}{0.08\linewidth}{
\begin{tabular}{cccccc}
\toprule
\textbf{Class 10\%} & Train $\mathcal{D}_f(\downarrow)$ & Train $\mathcal{D}_r(\uparrow)$ & test $\mathcal{D}_f$ & test $\mathcal{D}_r(\uparrow)$ & $\mathbf{MIA}^e(\uparrow)$ \\
\midrule
Pretrained & 97.270 & 96.180 & 85.800 & 84.063 & 0.170 \\
Retrain & 70.990 & 94.025 & 70.600 & 83.419 & 0.683 \\
OPC (ours) & 33.000 & 98.481 & 27.600 & 80.929 & 1.000 \\
\midrule
RL\citep{RL} & 92.300 & 99.620 & 78.200 & 82.374 & 0.980 \\
FT\citep{FT} & 80.450 & 99.662 & 68.400 & 80.307 & 0.480 \\
SSD\citep{SSD} & 84.690 & 95.390 & 73.000 & 83.641 & 0.722 \\
SalUn\citep{salun} & 84.540 & 99.677 & 67.200 & 82.707 & 1.000 \\
\bottomrule
\end{tabular}%
}
\resizebox{!}{0.08\linewidth}{
\begin{tabular}{cccccc}
\toprule
\textbf{Element 10\%} & $\mathcal{D}_f(\downarrow)$ & $\mathcal{D}_r(\uparrow)$ & $\mathcal{D}_{test}(\uparrow)$ & $\mathbf{MIA}^e(\uparrow)$ \\
\midrule
Pretrained & 97.520 & 97.576 & 83.837 & 0.119 \\
Retrain & 86.440 & 98.506 & 85.437 & 0.298 \\
OPC (ours) & 16.070 & 96.579 & 77.836 & 0.980 \\
\midrule
RL\citep{RL} & 95.480 & 98.720 & 83.457 & 0.224 \\
FT\citep{FT} & 90.010 & 99.912 & 81.036 & 0.290 \\
SSD\citep{SSD} & 97.630 & 97.543 & 83.797 & 0.120 \\
SalUn\citep{salun} & 96.030 & 98.524 & 83.737 & 0.189 \\
\bottomrule
\end{tabular}%
}
\label{tab:tinyimagenet-cls10-featuremap}
\end{table*}

Consistent to CIFAR10 and SVHN results, the TinyImagenet results show that the recovery through FM-recovery was possible, indicating that the phenomenon of the shallow forgetting in MU is not limited on Resnet or small datasets. 

\subsection{Relearning attack performance}
In this section, we present the full tables of the relearning attack performances for the CIFAR10 and SVHN experiments across various unlearning scenarios. 

We follow \cite{relearning}'s relearning protocol which consider a small dataset consist of 500 samples (50 per each classes) from train dataset and train the unlearned model using it. For all baselines, we first used SGD optimizer with learning rate 0.01 and weight decay 1e-5; if SGD with this basic setting invoke catastrophic forgetting (RA collapse), we further perform with learning rate 0.1 and 0.001 and selected the best. 

The detailed results are provided in \cref{tab:relearnatk_cls30_cifar10_svhn}, \cref{tab:relearnatk_halfcls30_cifar10_svhn}, and \cref{tab:relearnatk_random10_cifar10_svhn}.

\begin{table*}[h]
\caption{Relearning attack performance on 30\% Class unlearning scenario}
\centering
\resizebox{0.49\linewidth}{!}{%
\begin{tabular}{ccccccc}
\toprule
\textbf{CIFAR10} &
  \multicolumn{1}{c}{$\mathcal{D}_f (\downarrow)$} &
  \multicolumn{1}{c}{$\mathcal{D}_r (\uparrow)$} &
  \multicolumn{1}{c}{$\mathcal{D}_{test}^f(\downarrow)$} &
  \multicolumn{1}{c}{$\mathcal{D}_{test}^r (\uparrow)$} &
  \multicolumn{1}{c}{$\mathbf{MIA}^e (\uparrow)$} \\
\midrule
Pretrained (Original) & 99.444 & 99.416 & 94.800 & 94.400 & 0.015 \\
Retrain                          & 61.985          & 92.225 & 61.133          & 80.871 & 0.836          \\
\textbf{OPC} (ours)              & \textbf{36.348} & 99.822 & \textbf{32.967} & 93.557 & \textbf{1.000} \\ \midrule
GA \cite{GA}                     & 93.682          & 99.022 & 84.767          & 94.071 & 0.167          \\
RL \cite{RL}                     & 85.163          & 99.184 & 79.800          & 93.286 & 0.688          \\
BE \cite{BEBS}                   & 99.556          & 99.298 & 96.200          & 93.200 & 0.017          \\
FT \cite{FT}                     & 86.578          & 98.879 & 83.800          & 92.571 & 0.379          \\
NGD \cite{NGD}                   & 86.644          & 98.683 & 83.233          & 92.886 & 0.367          \\
NegGrad+ \cite{SCRUB_NegGrad+}   & 82.926          & 97.794 & 78.433          & 92.586 & 0.514          \\
EUk \cite{EUk/CFk}               & 87.059          & 89.086 & 82.600          & 84.014 & 0.266          \\
CFk \cite{EUk/CFk}               & 94.437          & 99.619 & 87.167          & 94.100 & 0.319          \\
SalUn \cite{salun}               & 88.059          & 99.683 & 82.267          & 94.200 & 0.589          \\
SCRUB \cite{SCRUB_NegGrad+}      & 79.341          & 97.749 & 73.233          & 92.486 & 0.407          \\
BT \cite{bad_teacher}            & 99.504          & 99.083 & 96.200          & 92.314 & 0.159          \\
$l1$-sparse \cite{l1sparse/MIAp} & 94.407          & 99.349 & 88.733          & 93.657 & 0.186          \\
MUNBa \cite{munba}               & 94.993          & 99.492 & 88.367          & 93.886 & 0.195          \\
DELETE \cite{delete}             & 99.474          & 99.257 & 95.133          & 93.171 & 0.046         
\\
\bottomrule
\end{tabular}%
}
\resizebox{0.49\linewidth}{!}{%
\begin{tabular}{ccccccc}
\toprule
\textbf{SVHN} &
  \multicolumn{1}{c}{$\mathcal{D}_f (\downarrow)$} &
  \multicolumn{1}{c}{$\mathcal{D}_r (\uparrow)$} &
  \multicolumn{1}{c}{$\mathcal{D}_{test}^f(\downarrow)$} &
  \multicolumn{1}{c}{$\mathcal{D}_{test}^r (\uparrow)$} &
  \multicolumn{1}{c}{$\mathbf{MIA}^e (\uparrow)$} \\
\midrule
Pretrained (Original) & 99.531 & 99.172 & 94.960 & 91.110 & 0.009 \\
Retrain                          & 61.309          & 89.066 & 55.750          & 81.416 & 0.947          \\
\textbf{OPC} (ours)              & \textbf{64.401} & 93.290 & \textbf{61.117} & 86.290 & \textbf{1.000} \\ \midrule
GA \cite{GA}                     & 89.394          & 91.072 & 83.333          & 84.242 & 0.180          \\
RL \cite{RL}                     & 78.882          & 94.245 & 74.391          & 86.669 & 0.920          \\
BE \cite{BEBS}                   & 88.475          & 90.902 & 82.051          & 84.355 & 0.217          \\
FT \cite{FT}                     & 82.242          & 95.357 & 79.967          & 87.932 & 0.343          \\
NGD \cite{NGD}                   & 83.459          & 95.813 & 78.857          & 88.444 & 0.339          \\
NegGrad+ \cite{SCRUB_NegGrad+}   & 37.372          & 2.295  & 38.310          & 2.015  & 0.132          \\
EUk \cite{EUk/CFk}               & 66.881          & 85.922 & 64.174          & 81.749 & 0.890          \\
CFk \cite{EUk/CFk}               & 79.804          & 88.531 & 74.955          & 82.926 & 0.354          \\
SalUn \cite{salun}               & 79.816          & 93.883 & 75.455          & 87.666 & 0.828          \\
SCRUB \cite{SCRUB_NegGrad+}      & 65.773          & 79.423 & 59.398          & 76.217 & 0.316          \\
BT \cite{bad_teacher}            & 83.331          & 91.545 & 80.777          & 84.535 & 0.864          \\
$l1$-sparse \cite{l1sparse/MIAp} & 79.740          & 94.025 & 76.074          & 88.032 & 0.382          \\
MUNBa \cite{munba}               & 88.230          & 93.627 & 84.789          & 86.071 & 0.348          \\
DELETE \cite{delete}             & 81.362          & 89.995 & 78.703          & 83.637 & 0.408         
\\
\bottomrule
\end{tabular}
}
\label{tab:relearnatk_cls30_cifar10_svhn}
\end{table*}

\begin{table*}[h]
\caption{Relearning attack performance on Half Class 30\% unlearning scenario. $\mathcal{D}_r^e$ and $\mathcal{D}_{test}^f$ represents the entangled retain set and test set with class labels 0,1 and 2. $\mathcal{D}_r^r$ and $\mathcal{D}_{test}^r$ are the same but with labels 3-9.}
\centering
\resizebox{0.49\linewidth}{!}{%
\begin{tabular}{ccccccc}
\toprule
\textbf{CIFAR10} & $\mathcal{D}_f (\downarrow)$ & $\mathcal{D}_r^e(\uparrow)$ & $\mathcal{D}_r^r(\uparrow)$ & $\mathcal{D}_{test}^f(\uparrow)$ & $\mathcal{D}_{test}^r(\uparrow)$ & $\mathbf{MIA}^e(\uparrow)$ \\
\midrule
Pretrained   (Original)             & 99.570           & 99.416           & 99.399           & 94.800              & 94.400              & 0.012          \\
Retrain                          & 86.282 & 99.902 & 99.901 & 84.667          & 86.429          & 0.663 \\
\textbf{OPC} (ours) &
  \textbf{62.089} &
  \textbf{99.987} &
  \textbf{99.961} &
  87.733 &
  93.171 &
  \textbf{0.993} \\ \midrule
\ GA \cite{GA}                   & 93.704 & 99.286 & 98.295 & 84.433          & 94.057          & 0.170 \\
RL \cite{RL}                     & 88.607 & 99.819 & 99.689 & 77.800          & 89.000          & 0.280 \\
BE \cite{BEBS}                   & 99.526 & 99.168 & 99.169 & 96.133 & 92.171          & 0.620 \\
FT \cite{FT}                     & 85.704 & 98.029 & 97.341 & 82.267          & 90.971          & 0.268 \\
NGD \cite{NGD}                   & 83.970 & 99.098 & 98.622 & 82.200          & 90.771          & 0.266 \\
NegGrad+ \cite{SCRUB_NegGrad+}   & 88.667 & 99.806 & 99.621 & 84.000          & 92.986          & 0.192 \\
EUk \cite{EUk/CFk}               & 92.474 & 99.657 & 98.607 & 86.000          & 93.457          & 0.138 \\
CFk \cite{EUk/CFk}               & 97.037 & 99.949 & 99.807 & 88.833          & 94.500 & 0.057 \\
SalUn \cite{salun}               & 84.341 & 98.816 & 98.669 & 78.800          & 85.629          & 0.352 \\
SCRUB \cite{SCRUB_NegGrad+}      & 94.889 & 99.143 & 98.395 & 87.433          & 93.557          & 0.811 \\
BT \cite{bad_teacher}            & 84.178 & 99.483 & 99.454 & 89.667          & 93.886          & 0.961 \\
$l1$-sparse \cite{l1sparse/MIAp} & 87.511 & 98.733 & 98.528 & 85.700          & 90.843          & 0.248 \\
MUNBa \cite{munba}               & 87.244 & 97.251 & 96.382 & 82.800          & 90.471          & 0.282 \\
DELETE \cite{delete}             & 96.919 & 98.911 & 98.635 & 91.700          & 93.086          & 0.288 \\
\bottomrule
\end{tabular}%
}
\resizebox{0.49\linewidth}{!}{%
\begin{tabular}{ccccccc}
\toprule
\textbf{SVHN} & $\mathcal{D}_f (\downarrow)$ & $\mathcal{D}_r^e(\uparrow)$ & $\mathcal{D}_r^r(\uparrow)$ & $\mathcal{D}_{test}^f(\uparrow)$ & $\mathcal{D}_{test}^r(\uparrow)$ & $\mathbf{MIA}^e(\uparrow)$ \\
\midrule
Pretrained (Original)            & 99.535 & 99.172 & 99.278 & 94.960          & 91.110          & 0.008 \\
Retrain                          & 78.946 & 86.700 & 86.097 & 77.020          & 78.418          & 0.383 \\
\textbf{OPC} (ours) &
  \textbf{30.549} &
  \textbf{99.721} &
  \textbf{99.643} &
  67.549 &
  91.024 &
  \textbf{0.999} \\ \midrule
GA \cite{GA}                     & 91.510 & 97.700 & 96.380 & 83.151          & 88.278          & 0.158 \\
RL \cite{RL}                     & 89.893 & 93.404 & 92.935 & 87.200          & 88.398          & 0.365 \\
BE \cite{BEBS}                   & 88.887 & 97.366 & 95.293 & 76.137          & 86.715          & 0.311 \\
FT \cite{FT}                     & 90.702 & 97.890 & 97.520 & 86.709          & 91.656          & 0.219 \\
NGD \cite{NGD}                   & 89.598 & 97.688 & 97.259 & 86.891          & 92.487          & 0.229 \\
NegGrad+ \cite{SCRUB_NegGrad+}   & 65.755 & 95.223 & 89.856 & 60.599          & 88.790          & 0.402 \\
EUk \cite{EUk/CFk}               & 88.849 & 94.724 & 93.817 & 83.370          & 86.742          & 0.164 \\
CFk \cite{EUk/CFk}               & 91.458 & 95.157 & 94.819 & 85.917          & 87.447          & 0.144 \\
SalUn \cite{salun}               & 97.173 & 99.329 & 99.340 & 95.479          & 92.899          & 0.083 \\
SCRUB \cite{SCRUB_NegGrad+}      & 59.419 & 89.132 & 81.916 & 54.995 & 83.278          & 0.620 \\
BT \cite{bad_teacher}            & 64.749 & 98.977 & 99.105 & 85.353          & 90.685          & 0.992 \\
$l1$-sparse \cite{l1sparse/MIAp} & 89.636 & 98.420 & 98.143 & 88.419          & 90.585          & 0.185 \\
MUNBa \cite{munba}               & 91.609 & 96.791 & 96.475 & 90.420          & 93.923 & 0.191 \\
DELETE \cite{delete}             & 75.998 & 97.543 & 92.316 & 64.183          & 87.434          & 0.491 \\
\bottomrule
\end{tabular}
}
\label{tab:relearnatk_halfcls30_cifar10_svhn}
\end{table*}

\begin{table*}[h]
\caption{Relearning attack performance on 10\% random unlearning scenario}
\centering
\resizebox{0.49\linewidth}{!}{%
\begin{tabular}{ccccccc}
\toprule
\textbf{CIFAR10} & $\mathcal{D}_f (\downarrow)$ & $\mathcal{D}_r(\uparrow)$ & $\mathcal{D}_{test}(\uparrow)$ & $\mathbf{MIA}^e(\uparrow)$ & $\mathbf{MIA}^p(\downarrow)$ \\ \midrule
Pretrained (Original)            & 99.356 & 99.432 & 94.520 & 0.015 & 0.545 \\
Retrain                          & 90.489 & 99.956 & 90.110 & 0.575 & 0.148 \\
\textbf{OPC} (ours)              & 74.600 & 99.993 & 92.820 & 0.576 & 1.000 \\ \midrule
GA\citep{GA}                     & 99.289 & 99.432 & 94.420 & 0.545 & 0.017 \\
RL\citep{RL}                     & 93.800 & 99.963 & 93.770 & 0.571 & 0.277 \\
BE\citep{BEBS}                   & 99.378 & 99.432 & 94.510 & 0.546 & 0.015 \\
FT\citep{FT}                     & 95.467 & 99.761 & 93.160 & 0.547 & 0.082 \\
NGD\citep{NGD}                   & 95.489 & 99.751 & 93.630 & 0.546 & 0.081 \\
NegGrad+\citep{SCRUB_NegGrad+}   & 95.911 & 99.788 & 93.440 & 0.548 & 0.081 \\
EUk\citep{EUk/CFk}               & 99.067 & 99.857 & 93.720 & 0.540 & 0.016 \\
CFk\citep{EUk/CFk}               & 99.267 & 99.941 & 94.070 & 0.540 & 0.016 \\
SalUn\citep{salun}               & 93.444 & 99.941 & 93.900 & 0.570 & 0.278 \\
SCRUB\citep{SCRUB_NegGrad+}      & 99.244 & 99.504 & 94.180 & 0.550 & 0.046 \\
BT\citep{bad_teacher}            & 92.711 & 99.370 & 93.460 & 0.568 & 0.489 \\
$l1$-sparse\citep{l1sparse/MIAp} & 94.911 & 98.699 & 92.520 & 0.541 & 0.104 \\
MUNBa \citep{munba}              & 95.800 & 99.274 & 93.200 & 0.548 & 0.106 \\
DELETE \citep{delete}            & 99.333 & 99.368 & 94.010 & 0.546 & 0.028 \\
\bottomrule
\end{tabular}%
}
\resizebox{0.49\linewidth}{!}{%
\begin{tabular}{ccccccc}
\toprule
\textbf{SVHN} & $\mathcal{D}_f (\downarrow)$ & $\mathcal{D}_r(\uparrow)$ & $\mathcal{D}_{test}(\uparrow)$ & $\mathbf{MIA}^e(\uparrow)$ & $\mathbf{MIA}^p(\downarrow)$ \\ \midrule
Pretrained (Original)            & 99.151 & 99.334 & 92.736 & 0.015 & 0.563 \\
Retrain                          & 89.459 & 98.241 & 88.038 & 0.569 & 0.159 \\
\textbf{OPC} (ours)              & 15.956 & 99.926 & 92.171 & 0.582 & 1.000 \\ \midrule
GA\citep{GA}                     & 97.209 & 97.563 & 88.284 & 0.561 & 0.058 \\
RL\citep{RL}                     & 92.098 & 96.355 & 91.138 & 0.537 & 0.242 \\
BE\citep{BEBS}                   & 96.269 & 96.650 & 86.136 & 0.568 & 0.083 \\
FT\citep{FT}                     & 93.690 & 99.976 & 93.543 & 0.551 & 0.101 \\
NGD\citep{NGD}                   & 93.812 & 99.971 & 93.896 & 0.549 & 0.098 \\
NegGrad+\citep{SCRUB_NegGrad+}   & 93.690 & 99.904 & 92.682 & 0.556 & 0.106 \\
EUk\citep{EUk/CFk}               & 92.234 & 95.051 & 86.382 & 0.550 & 0.113 \\
CFk\citep{EUk/CFk}               & 96.587 & 97.691 & 87.734 & 0.563 & 0.076 \\
SalUn\citep{salun}               & 91.521 & 97.723 & 90.727 & 0.555 & 0.307 \\
SCRUB\citep{SCRUB_NegGrad+}      & 97.619 & 97.713 & 88.291 & 0.561 & 0.057 \\
BT\citep{bad_teacher}            & 91.415 & 98.834 & 89.855 & 0.587 & 0.494 \\
$l1$-sparse\citep{l1sparse/MIAp} & 91.810 & 97.594 & 89.897 & 0.555 & 0.153 \\
MUNBa \citep{munba}              & 93.266 & 99.980 & 93.696 & 0.576 & 0.223 \\
DELETE \citep{delete}            & 96.178 & 96.546 & 85.871 & 0.569 & 0.091 \\
\bottomrule
\end{tabular}
}
\label{tab:relearnatk_random10_cifar10_svhn}
\end{table*}

Consistent with the evaluations in \cref{subsec:recovery-FMrecovery} and \cref{subsec:OPC-FMR}, the results demonstrate that OPC exhibits robust resistance against relearning attacks while conventional methods were easily recovered by relearning. However, in random and half-unlearning, the unlearned performances were too bad except for OPC and BT, the difference driven by relearning is incremental.

Although we tuned the learning rate for all baselines (especially EUk and Neggrad+ was sensitive) the relearning could not preserve RA on SVHN class 30\% unlearning for Neggrad+, since relearning is highly unstable and sensitive to the hyperparameters. 

\clearpage
\section{Head recovery}\label{subsec:exp-headrecovery}

\begin{table*}[h]
\caption{Recovered performance with head recovery on 30\% Class unlearning scenario}
\resizebox{0.5\linewidth}{!}{%
\begin{tabular}{cccccc}
\toprule
\textbf{CIFAR10} & Train $\mathcal{D}_f(\downarrow)$ & Train $\mathcal{D}_r(\uparrow)$ & test $\mathcal{D}_f(\downarrow)$ & test $\mathcal{D}_r(\uparrow)$ & $\mathbf{MIA}^e(\uparrow)$ \\
\midrule
Pretrained & 99.607 & 99.571 & 95.067 & 94.114 & 0.082 \\
Retrain & 71.963 & 95.213 & 72.400 & 85.557 & 0.750 \\
OPC (ours) & 33.333 & 99.156 & 31.633 & 91.214 & 0.976 \\
\midrule
GA\cite{GA} & 87.096 & 95.305 & 82.400 & 89.871 & 0.413 \\
RL\cite{RL} & 94.207 & 98.679 & 89.333 & 92.071 & 0.246 \\
BE\cite{BEBS} & 99.607 & 99.444 & 94.600 & 93.429 & 0.099 \\
FT\cite{FT} & 90.556 & 98.270 & 87.933 & 91.686 & 0.427 \\
NGD\cite{NGD} & 89.881 & 98.013 & 87.067 & 92.043 & 0.444 \\
NegGrad+\cite{SCRUB_NegGrad+} & 86.889 & 97.559 & 84.667 & 90.700 & 0.538 \\
EUk\cite{EUk/CFk} & 96.830 & 99.422 & 91.333 & 93.100 & 0.454 \\
CFk\cite{EUk/CFk} & 98.644 & 99.800 & 92.867 & 93.829 & 0.292 \\
SalUn\cite{salun} & 95.956 & 99.406 & 91.500 & 93.200 & 0.208 \\
SCRUB\cite{SCRUB_NegGrad+} & 88.956 & 97.048 & 84.367 & 91.457 & 0.453 \\
BT\cite{bad_teacher} & 99.481 & 99.495 & 93.500 & 94.029 & 0.175 \\
$l1$-sparse\cite{l1sparse/MIAp} & 94.963 & 99.149 & 89.667 & 92.671 & 0.372  \\
MUNBa \cite{munba}& 97.363 & 99.289 & 91.567 & 93.171 & 0.192 \\
DELETE \cite{delete}& 99.511 & 99.521 & 94.967 & 94.214 & 0.109 \\
\bottomrule
\end{tabular}%
}
\resizebox{0.5\linewidth}{!}{%
\begin{tabular}{cccccc}
\toprule
\textbf{SVHN} & Train $\mathcal{D}_f(\downarrow)$ & Train $\mathcal{D}_r(\uparrow)$ & test $\mathcal{D}_f(\downarrow)$ & test $\mathcal{D}_r(\uparrow)$ & $\mathbf{MIA}^e(\uparrow)$ \\
\midrule
Pretrained & 99.675 & 99.255 & 95.506 & 90.598 & 0.086 \\
Retrain & 89.292 & 96.221 & 89.465 & 85.326 & 0.440 \\
OPC (ours) & 47.154 & 99.521 & 45.524 & 91.376 & 1.000 \\
\midrule
GA\cite{GA} & 99.572 & 99.124 & 94.733 & 90.386 & 0.129 \\
RL\cite{RL} & 92.153 & 97.627 & 90.775 & 90.386 & 0.353 \\
BE\cite{BEBS} & 98.851 & 98.825 & 94.041 & 87.666 & 0.230 \\
FT\cite{FT} & 94.803 & 98.065 & 94.241 & 90.339 & 0.339 \\
NGD\cite{NGD} & 94.606 & 97.604 & 94.023 & 90.412 & 0.351 \\
NegGrad+\cite{SCRUB_NegGrad+} & 93.877 & 96.254 & 93.559 & 90.765 & 0.350 \\
EUk\cite{EUk/CFk} & 95.808 & 98.376 & 93.604 & 88.883 & 0.376 \\
CFk\cite{EUk/CFk} & 98.632 & 99.321 & 94.778 & 89.834 & 0.264 \\
SalUn\cite{salun} & 92.338 & 97.432 & 91.366 & 90.472 & 0.353 \\
SCRUB\cite{SCRUB_NegGrad+} & 91.786 & 87.612 & 91.012 & 83.019 & 0.786 \\
BT\cite{bad_teacher} & 93.661 & 98.098 & 92.394 & 89.408 & 0.420 \\
$l1$-sparse\cite{l1sparse/MIAp} & 92.788 & 95.631 & 92.213 & 88.464 & 0.444  \\
MUNBa \cite{munba} & 95.869 & 99.017 & 93.805 & 90.672 & 0.292 \\
DELETE \cite{delete} & 96.738 & 97.201 & 94.314 & 88.610 & 0.220 \\
\bottomrule
\end{tabular}%
}
\label{tab:cifar10-cls30-headrecovery}
\end{table*}

\begin{table*}[h]
\caption{Recovered performance with head recovery on 10\% random unlearning scenario}
\resizebox{0.5\linewidth}{!}{%
\begin{tabular}{cccccc}
\toprule
\textbf{CIFAR10} & $\mathcal{D}_f(\downarrow)$ & $\mathcal{D}_r(\uparrow)$ & $\mathcal{D}_{test}(\uparrow)$ & $\mathbf{MIA}^e(\uparrow)$  \\
\midrule
Pretrained & 99.644 & 99.575 & 94.400 & 0.094 \\
Retrain & 90.578 & 99.704 & 89.120 & 0.332 \\
OPC (ours) & 10.578 & 99.993 & 92.790 & 1.000 \\
\midrule
GA\cite{GA} & 99.444 & 99.560 & 94.290 & 0.094 \\
RL\cite{RL} & 93.689 & 99.968 & 93.850 & 0.360 \\
BE\cite{BEBS} & 99.622 & 99.565 & 94.390 & 0.096 \\
FT\cite{FT}& 95.711 & 99.812 & 93.060 & 0.227 \\
NGD\cite{NGD} & 96.089 & 99.807 & 93.610 & 0.238 \\
NegGrad+\cite{SCRUB_NegGrad+} & 96.378 & 99.840 & 93.390 & 0.227 \\
EUk\cite{EUk/CFk} & 99.178 & 99.867 & 93.630 & 0.152 \\
CFk\cite{EUk/CFk} & 99.422 & 99.956 & 94.150 & 0.114 \\
SalUn\cite{salun} & 93.689 & 99.963 & 93.920 & 0.342 \\
SCRUB\cite{SCRUB_NegGrad+} & 99.400 & 99.627 & 94.130 & 0.103 \\
BT\cite{bad_teacher} & 92.089 & 99.435 & 93.180 & 0.377 \\
$l1$-sparse\cite{l1sparse/MIAp} & 93.933 & 98.358 & 91.960 & 0.200 \\
MUNBa \cite{munba}& 95.289 & 99.022 & 92.720 & 0.198 \\
DELETE \cite{delete}& 99.444 & 99.533 & 94.110 & 0.109 \\
\bottomrule
\end{tabular}%
}
\resizebox{0.5\linewidth}{!}{%
\begin{tabular}{cccccc}
\toprule
\textbf{SVHN} & $\mathcal{D}_f(\downarrow)$ & $\mathcal{D}_r(\uparrow)$ & $\mathcal{D}_{test}(\uparrow)$ & $\mathbf{MIA}^e(\uparrow)$ \\
\midrule
Pretrained & 99.287 & 99.441 & 92.663 & 0.149 \\
Retrain & 92.765 & 99.998 & 92.033 & 0.271 \\
OPC (ours) & 40.983 & 99.933 & 92.371 & 1.000 \\
\midrule
GA\cite{GA} & 98.908 & 99.385 & 92.244 & 0.153 \\
RL\cite{RL} & 91.506 & 95.713 & 91.000 & 0.405 \\
BE\cite{BEBS} & 99.257 & 99.405 & 91.887 & 0.169 \\
FT\cite{FT} & 94.267 & 99.988 & 94.353 & 0.213 \\
NGD\cite{NGD} & 94.616 & 99.992 & 94.472 & 0.213 \\
NegGrad+\cite{SCRUB_NegGrad+} & 94.130 & 99.981 & 93.665 & 0.248 \\
EUk\cite{EUk/CFk} & 97.877 & 99.990 & 92.179 & 0.196 \\
CFk\cite{EUk/CFk} & 99.302 & 99.990 & 92.406 & 0.173 \\
SalUn\cite{salun} & 91.066 & 97.481 & 90.731 & 0.429 \\
SCRUB\cite{SCRUB_NegGrad+} & 99.257 & 99.508 & 92.628 & 0.148 \\
BT\cite{bad_teacher} & 93.159 & 98.773 & 90.988 & 0.566 \\
$l1$-sparse\cite{l1sparse/MIAp} & 93.523 & 98.970 & 92.601 & 0.279 \\
MUNBa\cite{munba} & 93.554 & 99.993 & 93.923 & 0.420 \\
DELETE\cite{delete} & 99.151 & 99.373 & 91.315 & 0.180 \\
\bottomrule
\end{tabular}%
}
\label{tab:cifar10-ele10-headrecovery}
\end{table*}

\begin{table*}[h]
\caption{Recovered performance with head recovery on 30\% Half Class unlearning scenario. $\mathcal{D}_r^e$ and $\mathcal{D}_{test}^f$ represents the entangled retain set and test set with class labels 0,1 and 2. $\mathcal{D}_r^r$ and $\mathcal{D}_{test}^r$ are the same but with labels 3-9.}
\centering
\resizebox{0.49\linewidth}{!}{%
\begin{tabular}{ccccccc}
\toprule
CIFAR10 & $\mathcal{D}_f (\downarrow)$ & $\mathcal{D}_r^e(\uparrow)$ & $\mathcal{D}_r^r(\uparrow)$ & $\mathcal{D}_{test}^f(\uparrow)$ & $\mathcal{D}_{test}^r(\uparrow)$ & $\mathbf{MIA}^e(\uparrow)$ \\
\midrule
Pretrained   (Original) & 99.689 & 99.526 & 99.571 & 95.067 & 94.114 & 0.080 \\
Retrain & 88.104 & 100.000 & 100.000 & 87.067 & 87.571 & 0.410 \\
\textbf{OPC}   (ours) & 33.289 & 99.659 & 100.000 & 78.533 & 93.229 & 1.000 \\
GA   \cite{GA} & 82.267 & 83.511 & 93.352 & 78.867 & 87.957 & 0.430 \\
RL   \cite{RL} & 93.170 & 99.956 & 99.854 & 86.333 & 87.957 & 0.272 \\
BE   \cite{BEBS} & 97.526 & 96.652 & 97.327 & 91.367 & 91.343 & 0.138 \\
FT   \cite{FT} & 92.133 & 97.867 & 98.019 & 89.067 & 90.429 & 0.270 \\
NGD   \cite{NGD} & 92.430 & 99.689 & 99.118 & 90.500 & 90.171 & 0.293 \\
NegGrad+   \cite{SCRUB_NegGrad+} & 95.704 & 99.941 & 99.803 & 92.167 & 92.100 & 0.275 \\
EUk   \cite{EUk/CFk} & 89.422 & 91.304 & 85.051 & 83.900 & 79.629 & 0.324 \\
CFk   \cite{EUk/CFk} & 98.519 & 99.985 & 99.991 & 93.233 & 92.871 & 0.135 \\
SalUn   \cite{salun} & 87.911 & 98.726 & 99.143 & 83.433 & 84.943 & 0.467 \\
SCRUB   \cite{SCRUB_NegGrad+} & 95.156 & 95.822 & 98.340 & 89.733 & 92.314 & 0.357 \\
BT   \cite{bad_teacher} & 94.696 & 99.467 & 99.530 & 91.400 & 93.857 & 0.797 \\
$l1$-sparse   \cite{l1sparse/MIAp} & 90.637 & 98.726 & 98.483 & 88.100 & 89.886 & 0.269 \\
MUNBa   \cite{munba} & 94.104 & 97.067 & 97.041 & 90.300 & 88.557 & 0.212 \\
DELETE   \cite{delete} & 98.682 & 99.185 & 99.429 & 94.300 & 93.586 & 0.126 \\
\bottomrule
\end{tabular}%
}
\resizebox{0.48\linewidth}{!}{%
\begin{tabular}{ccccccc}
\toprule
SVHN & $\mathcal{D}_f (\downarrow)$ & $\mathcal{D}_r^e(\uparrow)$ & $\mathcal{D}_r^r(\uparrow)$ & $\mathcal{D}_{test}^f(\uparrow)$ & $\mathcal{D}_{test}^r(\uparrow)$ & $\mathbf{MIA}^e(\uparrow)$ \\
\midrule
Pretrained   (Original) & 99.690 & 99.660 & 99.258 & 95.506 & 90.585 & 0.102 \\
Retrain & 91.458 & 99.206 & 97.834 & 90.357 & 82.420 & 0.259 \\
\textbf{OPC}   (ours) & 47.475 & 99.660 & 99.886 & 72.289 & 90.685 & 1.000 \\
GA   \cite{GA} & 98.957 & 99.146 & 98.726 & 94.205 & 89.894 & 0.167 \\
RL   \cite{RL} & 94.489 & 95.578 & 92.138 & 93.450 & 87.094 & 0.306 \\
BE   \cite{BEBS} & 98.874 & 98.775 & 98.817 & 94.105 & 87.567 & 0.203 \\
FT   \cite{FT} & 95.971 & 99.206 & 97.249 & 95.169 & 90.745 & 0.268 \\
NGD   \cite{NGD} & 96.016 & 99.123 & 97.100 & 95.515 & 90.957 & 0.277 \\
NegGrad+   \cite{SCRUB_NegGrad+} & 96.198 & 99.478 & 98.252 & 95.706 & 91.981 & 0.149 \\
EUk   \cite{EUk/CFk} & 97.551 & 99.856 & 99.643 & 93.395 & 89.847 & 0.133 \\
CFk   \cite{EUk/CFk} & 98.458 & 99.713 & 99.131 & 94.441 & 90.173 & 0.141 \\
SalUn   \cite{salun} & 96.712 & 99.596& 99.772 & 95.7697&93.1981 & 0.135 \\
SCRUB   \cite{SCRUB_NegGrad+} & 93.000 & 93.121 & 91.553 & 91.257 & 86.516 & 0.418 \\
BT   \cite{bad_teacher} & 95.154 & 98.783 & 98.949 & 94.824 & 89.501 & 0.424 \\
$l1$-sparse   \cite{l1sparse/MIAp} & 95.812 & 99.894 & 99.392 & 95.197 & 91.310 & 0.236 \\
MUNBa   \cite{munba} & 95.744 & 98.390 & 96.885 & 95.924 & 93.524 & 0.199 \\
DELETE   \cite{delete} & 98.753 & 98.859 & 98.837 & 94.378 & 88.936 & 0.207 \\ 
\bottomrule
\end{tabular}%
}
\label{tab:cifar10-cls30ele50-headrecovery}
\end{table*}

\begin{table*}[h]
\centering
\caption{Recovered performance with head recovery on TinyImageNet}
\resizebox{!}{0.08\linewidth}{%
\begin{tabular}{cccccc}
\toprule
\textbf{Class 10\%} & Train $\mathcal{D}_f(\downarrow)$ & Train $\mathcal{D}_r(\uparrow)$ & test $\mathcal{D}_f(\downarrow)$ & test $\mathcal{D}_r(\uparrow)$ & $\mathbf{MIA}^e(\uparrow)$ \\
\midrule
Pretrained & 97.230 & 96.139 & 93.600 & 94.288 & 0.283 \\
Retrain & 70.720 & 94.082 & 92.000 & 93.888 & 0.756 \\
OPC (ours) & 31.820 & 98.459 & 36.800 & 93.265 & 1.000 \\
\midrule
RL\cite{RL} & 91.760 & 99.626 & 90.600 & 93.532 & 0.992 \\
FT\cite{FT} & 80.040 & 99.688 & 88.800 & 92.265 & 0.564 \\
SSD\cite{SSD} & 83.870 & 95.408 & 92.200 & 94.021 & 0.776 \\
SalUn\cite{salun} & 91.330 & 99.587 & 90.600 & 93.643 & 0.984 \\
\bottomrule
\end{tabular}%
}
\resizebox{!}{0.08\linewidth}{%
\begin{tabular}{cccccc}
\toprule
\textbf{Element 10\%} & $\mathcal{D}_f(\downarrow)$ & $\mathcal{D}_r(\uparrow)$ & $\mathcal{D}_{test}(\uparrow)$ & $\mathbf{MIA}^e(\uparrow)$ \\
\midrule
Pretrained & 96.230 & 96.296 & 84.237 & 0.303 \\
Retrain & 85.890 & 97.749 & 85.497 & 0.354 \\
OPC (ours) & 14.91 & 95.699 & 78.016 & 0.990 \\
\midrule
RL\cite{RL} & 93.250 & 97.533 & 83.497 & 0.542 \\
FT\cite{FT} & 88.930 & 99.576 & 81.076 & 0.335 \\
SSD\cite{SSD} & 96.180 & 96.211 & 83.957 & 0.286 \\
SalUn\cite{salun} & 94.270 & 97.448 & 83.497 & 0.492 \\
\bottomrule
\end{tabular}%
}
\label{tab:tinyimagenet-cls10-headrecovery}
\end{table*}

The remaining linear separability on unlearned forget features naturally indicates that the classification accuracy can be recovered with proper classifier head. 
In \cref{sec:recovery}, the composition of FM-recovery layer and classifier head of original model is the desired head which recovers the performance. 
In this section, we aim to try the same without the pretrained model, by mapping the unlearned features directly to the desired logits (the one-hot vector of target labels) with similar method. 

We consider following linear least square problem to find the recovered prediction head:

\begin{equation}
    W^*=\underset{W}{\arg\min} \sum_{(x,y)\in\mathcal{D}}\|Wf_{\theta^{un}}(x)-e_y\|_2^2 ,
\end{equation}
where $\mathcal{D}$ is a sample dataset, $\theta^{un}$ is the unlearned parameters and $e_y$ is the one-hot vector of label $y$ of sample $x$. 
We used $\mathcal{D}_{val}$ as sample dataset in implementation. For CIFAR10, we used normalized features instead of $f_{\theta^{un}}(x)$ since some models including retrained model lost performance on $\mathcal{D}_r$.

The results are shown in \cref{tab:cifar10-cls30-headrecovery}, \cref{tab:cifar10-ele10-headrecovery} for class 30\% unlearning, random 10\% unlearning experiment. Also, we include results on TinyImagenet result in \cref{tab:tinyimagenet-cls10-headrecovery}.
Similar to FM-recovery results, OPC and retraining are only methods which are resistant and all others were showing remarkable recovery, further supporting the residual linear separability and shallow forgetting.

\section{Unlearning inversion attack}\label{appendix:exp-inversion}

\begin{figure*}[h]
  \centering
  \begin{subfigure}[t]{0.99\textwidth}
      \includegraphics[width=\textwidth]{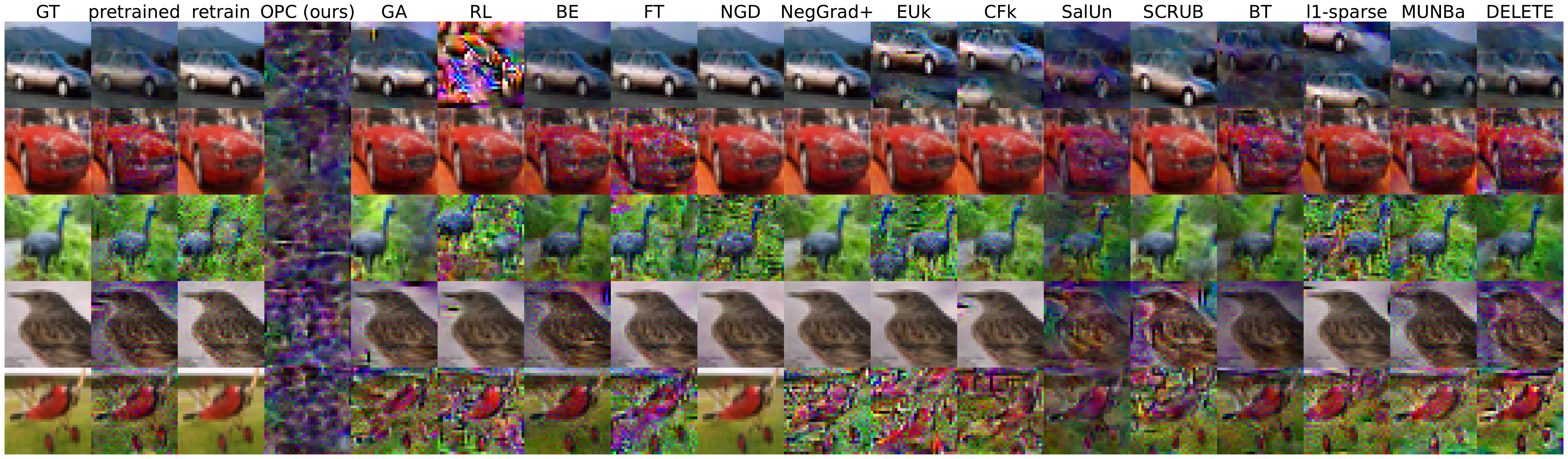}
      \caption{Reconstruction of forgotten images on CIFAR10 30\% class unlearning scenario}
      \label{subfig:inversion_cifar10cls30-detailed}
    \end{subfigure}
\begin{subfigure}[t]{0.99\textwidth}
      \includegraphics[width=\textwidth]{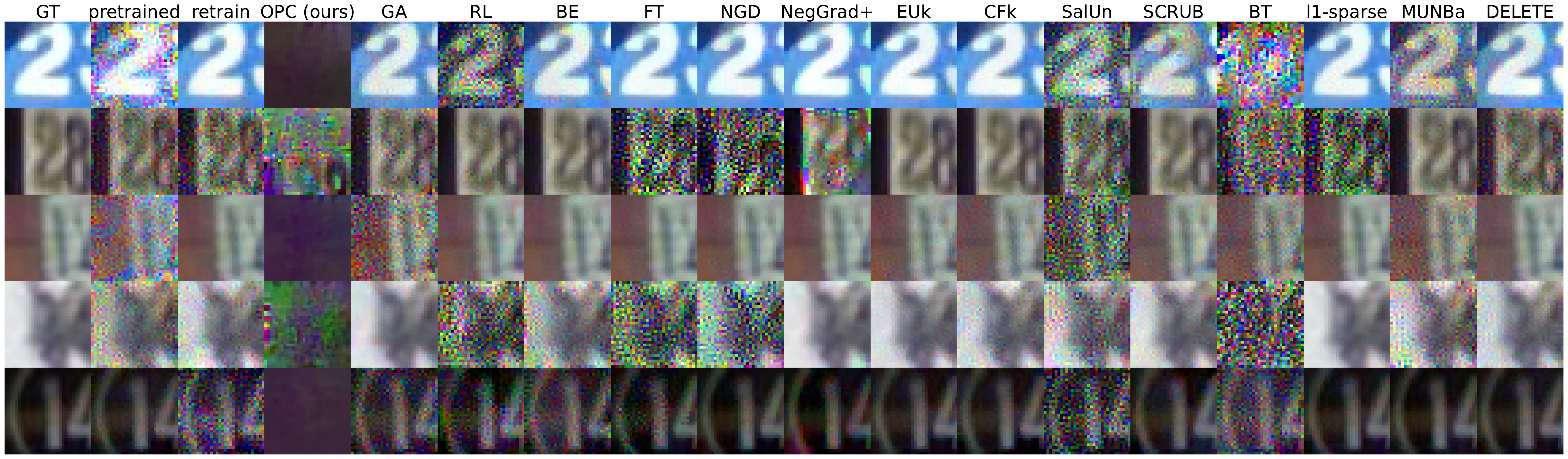}
      \caption{Reconstruction of forgotten images on SVHN 30\% class unlearning scenario}
      \label{subfig:inversion_svhncls30-detailed}
    \end{subfigure}

\caption{The results of unlearning inversion. The target images are sampled from the forget set $\mathcal{D}_f$ under 30\% class unlearning scenario. GT represents the ground truth image from the dataset and others are the results of inversion attacks from each unlearned model.}
\label{fig:inversion-recon-detailed}
\end{figure*}

\begin{figure*}[h]
  \centering
  \begin{subfigure}[t]{0.99\textwidth}
      \includegraphics[width=\textwidth]{plots/plots_inversion_reconstruct/cifar10_element10/FigF2a_inversion.png}
      \caption{Reconstruction of forgotten images on CIFAR10 10\% random unlearning scenario}
      \label{subfig:inversion_cifar10ele10}
    \end{subfigure}
   \begin{subfigure}[t]{0.99\textwidth}
      \includegraphics[width=\textwidth]{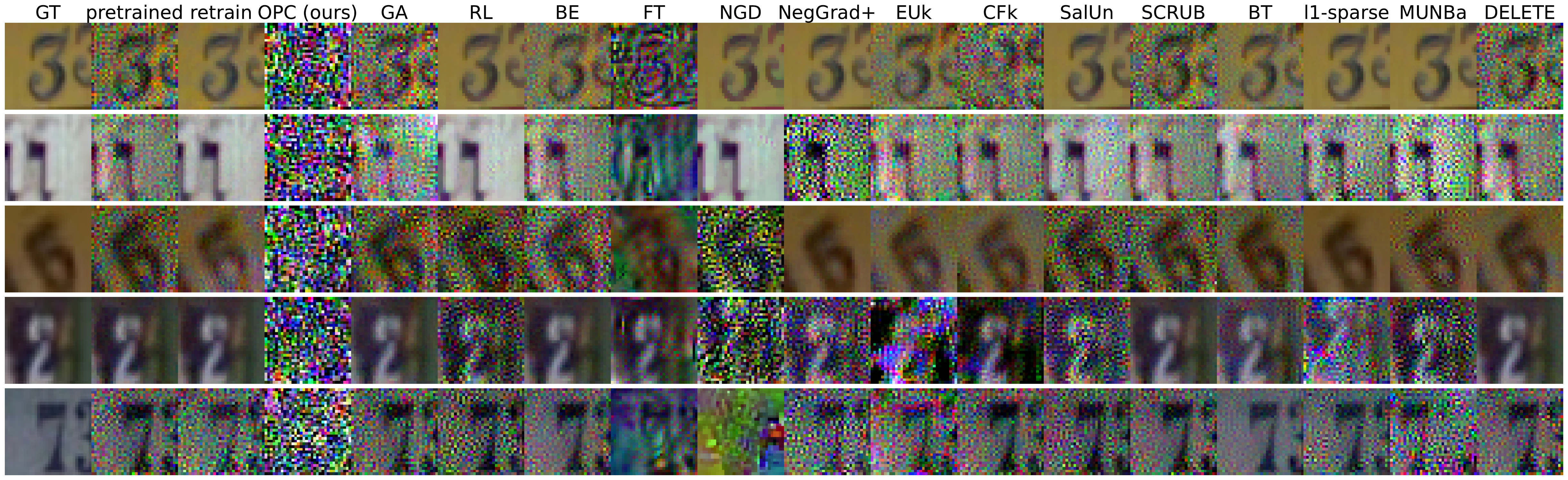}
      \caption{Reconstruction of forgotten images on SVHN 10\% random unlearning scenario}
      \label{subfig:inversion_svhnele10}
    \end{subfigure}
    \caption{The results of unlearning inversion. The target images are sampled from the forget set $\mathcal{D}_f$ under 10\% random unlearning scenario. GT represents the ground truth image from the dataset and others are the results of inversion attacks from each unlearned model.}
    \label{fig:inversion-recon-ele}
\end{figure*}

Recently, \cite{hu2024_reconattack} claimed the vulnerability of MU, with unlearning inversion attack, based on gradient-inversion, on unlearned model. Surprisingly, the attacker could reconstruct the sample image which were in the forget set $\mathcal{D}_f$. To visualize how the unlearning methods forget features, we exploit \cite{hu2024_reconattack}'s method and applied it to MU benchmarks and our method, to evaluate the vulnerability under unlearning inversion attack. 

Given sample image and corresponding label $(x,y)\in \mathcal{D}_f$ in forget set, the original \cite{hu2024_reconattack} implementation takes $\nabla^*$ as the parameter movement driven by unlearning process with single forget sample and find best sample $x'$ which makes $\nabla'(x')=\nabla_\theta \mathcal{L}_{CE}(f_\theta(x'),y)$ similar to $\nabla^*$, but unfortunately the unlearning problem setting does not meet theirs, since the forget set $\mathcal{D}_f$ is much larger compared to the single datapoint used in \cite{hu2024_reconattack}. Hence, we introduce an oracle providing true $\nabla_\theta \mathcal{L}_{CE}(f_{\theta}(x),y)$ as $\nabla^*$ for the reconstruction, which is quite strong advantage for the attacker and highly informative.

The results are collected in \cref{fig:inversion-recon-detailed} and \cref{fig:inversion-recon-ele}. 
Interestingly, almost all other unlearning methods including retrain were vulnerable under the inversion attack, while only our method \textbf{OPC} was consistently resistant. 

In random unlearning scenario with OPC, some forget images were recovered in CIFAR10, but this observation is may due to the imperfect unlearning, since the forget accuracy is still high (but much less than others) in \cref{tab:cifar10-ele10-perf}. 
The results on SVHN shows the high resistance of \textbf{OPC}, as the forgetting was extremely successful with significant gap on forget accuracy (7.5\% on \textbf{OPC}, $>90$ on others).

\clearpage
\section{Sensitivity analysis}

\subsection{Sensitivity on $\lambda_f$}\label{subsec:hparam}

\begin{table*}[h]
\caption{Hyperparameter analysis on CIFAR10, 30\% class unlearning scenario}
\centering
\begin{tabular}{ccccccc}
\toprule
$\mathcal{D}_f(\downarrow)$ & $\mathcal{D}_r(\uparrow)$ & test $\mathcal{D}_{f}(\uparrow)$ & test $\mathcal{D}_{r}(\uparrow)$ & $\mathbf{MIA}^e(\uparrow)$ &$\lambda_r$ & $\lambda_f$ \\ \midrule
2.037 & 99.737 & 2.333 & 94 & 1 & 1 & 1 \\
0.037 & 99.771 & 0.167 & 94.371 & 1 & 1 & 0.9 \\
1.592 & 99.603 & 1.867 & 93.757 & 1 & 1 & 0.8 \\
0 & 99.606 & 0 & 93.143 & 1 & 1 & 0.7 \\
3.4 & 99.759 & 3.333 & 93.543 & 1 & 1 & 0.6 \\
0.044 & 99.791 & 0 & 94.571 & 1 & 1 & 0.5 \\
1.793 & 99.851 & 1.733 & 94.457 & 1 & 1 & 0.4 \\
0.852 & 99.908 & 1.033 & 94.814 & 1 & 1 & 0.3 \\
1.696 & 99.810 & 1.5 & 94.7 & 1 & 1 & 0.2 \\
25.178 & 99.876 & 23.633 & 94.614 & 1 & 1 & 0.1 \\ \bottomrule
\end{tabular}
\label{tab:cifar10-cls30-hparam-analysis}
\end{table*}

\begin{table*}[h]
\caption{Hyperparameter analysis on CIFAR10, 10\% random unlearning scenario}
\centering
\begin{tabular}{ccccccc}
\toprule
 $\lambda_r$& $\lambda_f$ & $\mathcal{D}_f(\downarrow)$ & $\mathcal{D}_r(\uparrow)$ & $\mathcal{D}_{test}(\uparrow)$ & $\mathbf{MIA}^e(\uparrow)$ & $\mathbf{MIA}^p(\downarrow)$ \\
 \midrule
0.95 & 0.01 & 13.200 & 100 & 93.68 & 1 & 0.629 \\
0.95 & 0.03 & 12.578 & 100 & 93.19 & 1 & 0.625 \\
0.95 & 0.05 & 10.289 & 100 & 93.07 & 1 & 0.623 \\
0.95 & \textbf{0.07} & 8.26667 & 99.9926 & 92.92 & 1 & 0.625 \\
0.95 & 0.09 & 9.178 & 99.9951 & 92.95 & 1 & 0.634 \\
0.95 & 0.11 & 8.933 & 99.9753 & 92.86 & 1 & 0.636 \\
0.95 & 0.13 & 8.444 & 99.9704 & 92.73 & 1 & 0.630 \\
0.95 & 0.15 & 9.844 & 99.958 & 92.8 & 1 & 0.635 \\
0.95 & 0.17 & 7.978 & 99.963 & 92.48 & 1 & 0.630 \\
0.95 & 0.19 & 9.022 & 99.9481 & 92.15 & 1 & 0.633 \\ 
\bottomrule
\end{tabular}

\label{tab:cifar10-ele10-hparam-analysis}
\end{table*}

\cref{tab:cifar10-cls30-hparam-analysis} and \cref{tab:cifar10-ele10-hparam-analysis} present the sensitivity of OPC to the hyperparameters that control the relative contribution of the retain loss and the unlearning loss. Across both the class-unlearning and random-unlearning settings, OPC remains highly stable to the choice of coefficient $\lambda_f$. 

In the 30\% class-unlearning scenario, decreasing the unlearning coefficient shows that OPC maintains stable retain accuracy and effective forgetting across a broad range of values. However, when the coefficient becomes very small (e.g., 0.1), we observe that the training finishes before the contraction fully occurs, resulting in slightly higher residual forget accuracy. This suggests that extremely small coefficients may under-drive the contraction process.

A similar trend appears in the 10\% random-unlearning setting. The OPC forgetting was stable in $\lambda_f$ growth; however if $\lambda_f$ is too small then the contraction process may not be sufficient, leading to smaller UA.  

\subsection{Scaling analysis}\label{subsec:scale}

\begin{table*}[h]
\caption{Scaling analysis on CIFAR10}
\centering
\begin{tabular}{cccccccc}
\toprule
\multicolumn{8}{c}{Class unlearning scenario}\\ \midrule
Unit(10\%) & $\mathcal{D}_f(\downarrow)$ & $\mathcal{D}_r(\uparrow)$ & test $\mathcal{D}_{f}(\downarrow)$ & test $\mathcal{D}_{r}(\uparrow)$ & $\mathbf{MIA}^e(\uparrow)$ & $\lambda_r$ & $\lambda_f$ \\ \midrule
5 & 0 & 99.702 & 0 & 96.08 & 1 & 1 & 1 \\
4 & 0.006 & 99.426 & 0 & 94.417 & 1 & 1 & 0.9 \\
3 & 0 & 99.746 & 0 & 94.129 & 1 & 1 & 0.9 \\
2 & 0.089 & 99.606 & 0.1 & 93.4125 & 1 & 1 & 0.9 \\
1 & 0.022 & 99.412 & 0 & 93.167 & 1 & 1 & 0.9 \\ \bottomrule
\end{tabular}
\begin{tabular}{cccccccc}
\toprule
\multicolumn{8}{c}{Random unlearning scenario}\\ \midrule
$\mathcal{D}_f$ size & $\mathcal{D}_f(\downarrow)$ & $\mathcal{D}_r(\uparrow)$ & $\mathcal{D}_{test}(\uparrow)$ & $\mathbf{MIA}^e(\uparrow)$ & $\mathbf{MIA}^p(\downarrow)$ & $\lambda_r$ & $\lambda_f$ \\ \midrule
10\% & 10.978 & 99.948 & 92.060 & 0.649 & 1.000 & 1.000 & 0.150 \\
20\% & 10.700 & 99.981 & 91.880 & 0.719 & 1.000 & 1.000 & 0.150 \\
30\% & 10.585 & 99.991 & 88.410 & 0.793 & 1.000 & 1.000 & 0.150 \\
40\% & 11.306 & 99.985 & 85.950 & 0.836 & 1.000 & 1.000 & 0.150 \\
50\% & 11.618 & 99.991 & 78.640 & 0.881 & 1.000 & 1.000 & 0.150 \\
\bottomrule
\end{tabular}
\label{tab:cifar10-scale}
\end{table*}

In \cref{tab:cifar10-scale}, we present the results of applying OPC across various unlearning scenarios. In both the Class and Random unlearning settings, OPC maintains stable unlearning performance even as the size of the forget set increases, and this is achieved with simple hyperparameter adjustments.

Except for the hyperparameters explicitly shown, all other hyperparameters (with the exception of the number of epochs) follow the configuration in \cref{appendix:setup-details}. For the Class unlearning scenario, we use 25 epochs only when the forget ratio is 50\%, and 30 epochs for all other cases. For the Random unlearning scenario, we use 20 epochs for all experiments.

\section{Analysis of relation between OPC and BT\cite{bad_teacher}}\label{appendix:BT_analysis}

The BT\citep{bad_teacher} shares some behavioral similarities to OPC on feature visualization in \cref{fig:tsne-feature-halfcalss-svhn} on half-class unlearning scenario, that the forget features are showing disentanglement from retain features, with promising UA and $MIA^e$ scores. Also, BT sometimes show resistance against the reconstruction attack. 

We carefully hypothesize this partial similarity and success of BT is due to small-normed prediction, which induces high uncertainty by \cref{thm:entropyLB} and concept of OPC.

\begin{figure}[h]
    \centering
    \begin{subfigure}[t]{0.4\textwidth}
      \includegraphics[width=\textwidth]{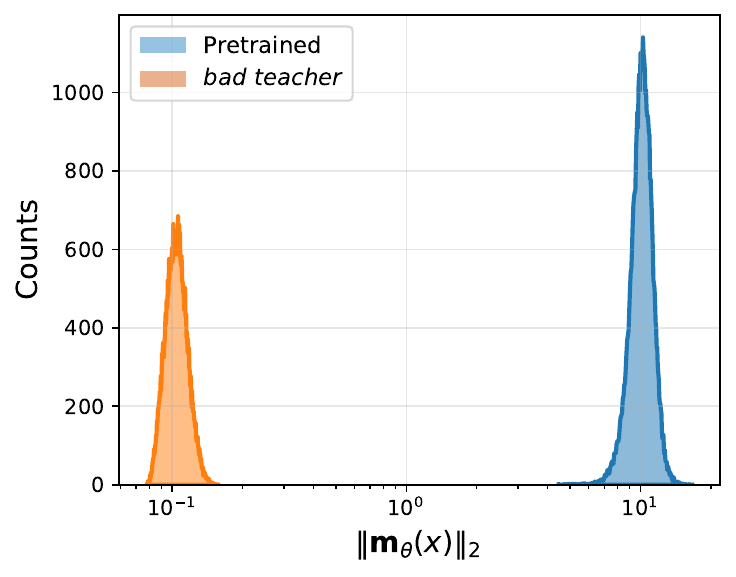}
      \caption{$\|\textbf{m}_\theta(x)\|_2$ on CIFAR10}
    \end{subfigure}
    \begin{subfigure}[t]{0.4\textwidth}
      \includegraphics[width=\textwidth]{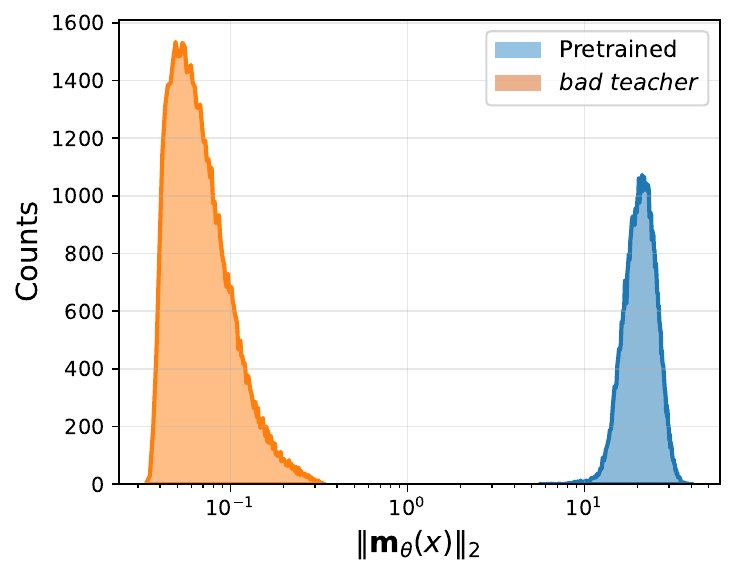}
      \caption{$\|\textbf{m}_\theta(x)\|_2$ on SVHN}
    \end{subfigure}
    \caption{Norm distribution of prediction of \textit{bad teacher} and pretrained model}
    \label{fig:norm-bad_teacher}
\end{figure}

Recall that BT employs a knowledge distillation from randomly initialized model, the bad teacher, to guide the broken prediction on forget dataset. Interestingly, the prediction norm from the bad teacher is consistently low compared to the pretrained model. 
As shown in \cref{fig:norm-bad_teacher} with full train dataset. The randomly initialized model gives prediction with much smaller (about 0.01 scale) norm. 

\begin{figure}[h]
    \centering
    \begin{subfigure}[t]{0.4\textwidth}
      \includegraphics[width=\textwidth]{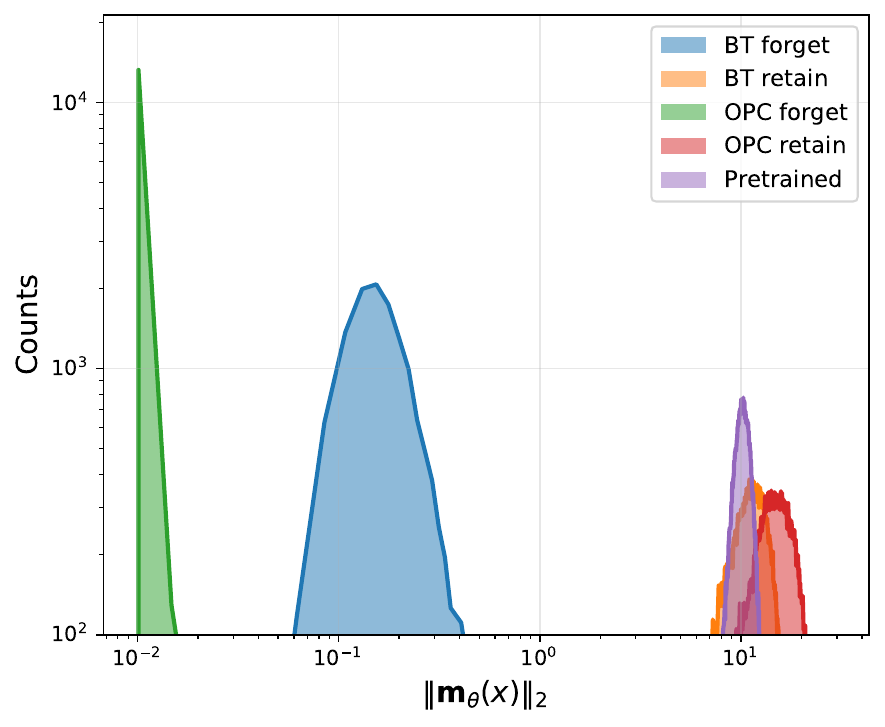}
      \caption{$\|\textbf{m}_\theta(x)\|_2$ on CIFAR10}
    \end{subfigure}
    \begin{subfigure}[t]{0.4\textwidth}
      \includegraphics[width=\textwidth]{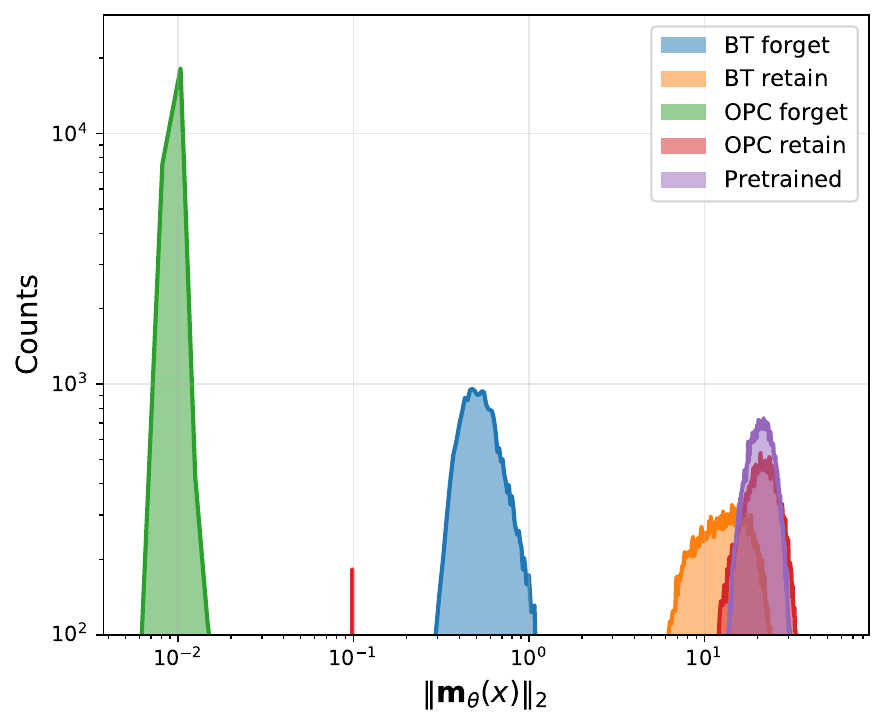}
      \caption{$\|\textbf{m}_\theta(x)\|_2$ on SVHN}
    \end{subfigure}
    \caption{Norm distribution of prediction of $\mathcal{D}_f$ and $\mathcal{D}_r$}
    \label{fig:norm-bad_teacher vs OPC vs Pretrained}
\end{figure}

Since BT-unlearned model is trained to imitate the teacher's behavior, the BT-unlearned model's prediction on forget set has small norm too, as depicted in \cref{fig:norm-bad_teacher vs OPC vs Pretrained}. Note that the prediction norm was preserved on retain set for both BT and OPC.

However, the magnitude degradation by BT is limited; there is a significant gap between OPC and BT on forget prediction norm, which exhibits the information destruction. 
Also, the BT training was insufficient to guide the feature-level forgetting via contraction, makes the unlearned model vulnerable under recovery or reconstruction attack.

\section{Limitations}\label{sec:limitations}

Our work demonstrates the shallow forgetting phenomenon of MU (machine unlearning) on classification benchmark, that they are easily reversible by \FMR{}: simple linear transformation, without use of train data. However, it requires to access the original model and therefore it is not an attack model but a unlearner-side evaluation. Regarding the feature correlation between independently trained model (usually explained and exploited in model merging literature), attacker can use their own model as a pretrained model, but this attack scenario is out of our research scope focusing on representation learning perspective.

For the proposed MU method OPC, we evaluated mainly on classification benchmark. While it can be used for generative model and performs well to invoke malfunction on forget data, naive OPC algorithm itself may not be the perfect solution if user wants the model to generate realistic output but remove the concept only; we have modification for this scenario and confirmed the efficacy on UnlearnCanvas benchmark, but we discarded from this manuscript, to focus on current research focus. 

Some of OPC's benefits stay for the classification task; especially about the entropy explosion and gradient collapse; the theory focuses on the classification and cross-entropy loss function. The concept of "removing distinguishability" survives for any task on deep learning, but some of evaluations would not work outside of the classification. 

\section{OPC for generative model unlearning}\label{subsec:OPC-DDPMnaive}

The core idea of OPC, collapsing the model prediction to a single point (the origin), is not limited to classification models but extends to various types of representation learning.

Recall from the results of generative decoder in \cref{subsec:recovery-decoder} that the training dynamics of minimizing \cref{eq:opc_loss_ce} facilitate the selective removal of information from forget features.

In this section, we aim to unlearn a DDPM model trained on CIFAR10 to generate images conditioned on class embedding vectors; thereby, evaluating how OPC generalizes to generative models.

For implementation, we consider the class embedding module of the model as $f_\theta$ and replace the cross-entropy loss in \cref{eq:opc_loss_ce} with the DDPM loss. 
In contrast to classification models, we apply OPC loss directly to features, as the model architecture does not include a prediction head. 
The modified loss function is defined as:

\begin{equation}
    \begin{aligned}
        &\mathcal{L}_{OPC}^{DDPM}=\mathbb{E}_{(x_0,c)\sim\mathcal{D}_f}\|f_\theta(c)\|_2 \\
        &+\mathbb{E}_{(x_0,c)\sim\mathcal{D}_r,t,\epsilon} \|\epsilon-\epsilon_{\theta}(\sqrt{\bar{\alpha_t}}x_0+\sqrt{1-\bar{\alpha_t}}\epsilon,f_\theta(c),t)\|_2^2
    \end{aligned}
\end{equation}

where $c$ represents the class label of image. In our experiments, we focus on unlearning a single class, ``airplane" (class label 0), from the pre-trained DDPM. After performing MU, we use an external classifier to classify and evaluate the generated images. 

\begin{table}[h]
    \centering
    \caption{Quantitative results of DDPM unlearning experiment performed on CIFAR10 10\% class unlearning.}\label{tab:ddpm}
    \resizebox{0.4\linewidth}{!}{
    \begin{tabular}{ccc}
    \toprule
    \midrule
    \textbf{Methods} & \textbf{UA} ($\uparrow$)   & \textbf{FID} ($\downarrow$)\\
    \midrule
    Original & 3.60 & 15.67  \\
    Retrain & 99.97 & 13.49 \\
    \midrule
    SalUn \citep{salun} & 99.99 & 17.33 \\
    \textbf{OPC} (ours) & 99.98 & 16.06 \\
    \midrule
    \bottomrule
    \end{tabular}
    }
    \label{tab: ddpm}
    \vspace{-0.3cm}
\end{table}

\begin{table*}[t]
    \centering
    \caption{Comparison of Images generated from unlearned DDPM, trained on CIFAR10. The unlearning was performed to forget single class 0, the airplane.}
    \resizebox{\textwidth}{!}{
    \begin{tabular}{ccccc|ccccccccc}
  \toprule
   \multirow{2}{*}{\textbf{Methods}}
  & \multicolumn{4}{c}{\textbf{Forgetting class: `Airplane'}} 
  & \multicolumn{9}{c}{\textbf{Retain classes}} \\
    & I1 & I2 & I3 & I4
    & automobile & bird & cat & deer & dog &frog & horse & ship & truck \\
  \midrule
{\raisebox{0.04\linewidth}{SalUn}}  &
\includegraphics[width=0.12\linewidth]{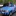} &
\includegraphics[width=0.12\linewidth]{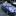} &
\includegraphics[width=0.12\linewidth]{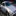} &
\includegraphics[width=0.12\linewidth]{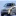} &
\includegraphics[width=0.12\linewidth]{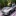} &
\includegraphics[width=0.12\linewidth]{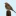} &
\includegraphics[width=0.12\linewidth]{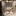} &
\includegraphics[width=0.12\linewidth]{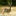} &
\includegraphics[width=0.12\linewidth]{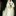} &
\includegraphics[width=0.12\linewidth]{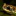} &
\includegraphics[width=0.12\linewidth]{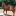} &
\includegraphics[width=0.12\linewidth]{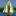} &
\includegraphics[width=0.12\linewidth]{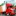} \\
  \midrule
{\raisebox{0.04\linewidth}{\textbf{OPC}}}  &
\includegraphics[width=0.12\linewidth]{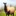} &
\includegraphics[width=0.12\linewidth]{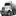} &
\includegraphics[width=0.12\linewidth]{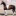} &
\includegraphics[width=0.12\linewidth]{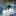} &
\includegraphics[width=0.12\linewidth]{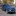} &
\includegraphics[width=0.12\linewidth]{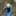} &
\includegraphics[width=0.12\linewidth]{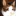} &
\includegraphics[width=0.12\linewidth]{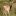} &
\includegraphics[width=0.12\linewidth]{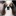} &
\includegraphics[width=0.12\linewidth]{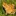} &
\includegraphics[width=0.12\linewidth]{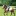} &
\includegraphics[width=0.12\linewidth]{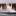} &
\includegraphics[width=0.12\linewidth]{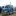} \\
    \midrule
    \bottomrule[1pt]
\end{tabular}
}
    \label{fig:ddpm-qualitative}
\end{table*}

The results are presented in \cref{tab: ddpm}. Consistent with the results for classification models, OPC successfully unlearns the target class, achieving high UA. Although we push the embedding of the forget class toward 0, the denoising model generates high-fidelity images from OPC-unlearned class embedding, as the FID score in \cref{tab: ddpm} remains favorable.

\end{document}